\documentclass[journal,twoside,web]{ieeecolor}
\usepackage{generic}
\usepackage{cite}
\usepackage{amsmath,amssymb,amsfonts}
\usepackage{algorithmic}
\usepackage{graphicx}
\usepackage{textcomp}

\usepackage[ruled,vlined]{algorithm2e}

\usepackage{subfig}
\usepackage{multirow}

\usepackage{changepage}

\usepackage{caption}
\usepackage{array}
\newcolumntype{?}{!{\vrule width 1pt}}
\usepackage[hyphens]{url}

\usepackage{multicol}

\def\BibTeX{{\rm B\kern-.05em{\sc i\kern-.025em b}\kern-.08em
    T\kern-.1667em\lower.7ex\hbox{E}\kern-.125emX}}
\begin{document}
	
\twocolumn[  
\begin{@twocolumnfalse}
This article has been accepted for publication. Citation information: DOI 10.1109/TMI.2018.2872031, IEEE Transactions on Medical Imaging. See https://ieeexplore.ieee.org/document/8471199 for the published article.
	
\textcopyright 2018 IEEE. Personal use of this material is permitted. Permission from IEEE must be obtained for all other uses, in any current or future media, including reprinting/republishing this material for advertising or promotional purposes, creating new collective works, for resale or redistribution to servers or lists, or reuse of any copyrighted component of this work in other works.
		
\end{@twocolumnfalse}
]
	
\title{Joint Weakly and Semi-Supervised Deep Learning for Localization and Classification of Masses in Breast Ultrasound Images}
\author{Seung Yeon Shin, Soochahn Lee$^*$, \IEEEmembership{Member, IEEE}, Il Dong Yun, \IEEEmembership{Member, IEEE},\\ Sun Mi Kim, and Kyoung Mu Lee, \IEEEmembership{Senior Member, IEEE}
\thanks{S. Y. Shin and K. M. Lee are with 
the Department of Electrical and Computer Engineering, Automation and Systems Research Institute, Seoul National
University, 1 Gwanak-ro, Gwanak-gu, Seoul, 08826, South Korea (e-mail: syshin@snu.ac.kr, kyoungmu@snu.ac.kr).}
\thanks{S. Lee is with
the Department of Electronic Engineering, Soonchunhyang University, Asan, 31538, South Korea (e-mail: sclsch@sch.ac.kr). *\emph{Corresponding author.}}
\thanks{I. D. Yun is with
the Division of Computer and Electronic Systems Engineering, Hankuk University of Foreign Studies, Yongin, 17035, South Korea (e-mail: yun@hufs.ac.kr).}
\thanks{S. M. Kim is with
the Department of Radiology, Seoul National University Bundang Hospital, Seongnam, 13620, South Korea (e-mail: kimsmlms@daum.net).}
\thanks{\scriptsize{This work was supported by the National Research Foundation of Korea (NRF) grants funded by the Korean government (MSIT) (No. 2017R1A2B2011862, 2015R1C1A1A01054697, and 2017R1A2B4004503).}}
}

\maketitle

\begin{abstract}
We propose a framework for localization and classification of masses in breast ultrasound (BUS) images. We have experimentally found that training convolutional neural network based mass detectors with large, weakly annotated datasets presents a non-trivial problem, while overfitting may occur with those trained with small, strongly annotated datasets. To overcome these problems, we use a weakly annotated dataset together with a smaller strongly annotated dataset in a hybrid manner. We propose a systematic weakly and semi-supervised training scenario with appropriate training loss selection. Experimental results show that the proposed method can successfully localize and classify masses with less annotation effort. The results trained with only 10 strongly annotated images along with weakly annotated images were comparable to results trained from 800 strongly annotated images, with the 95\% confidence interval of difference -3.00\%--5.00\%, in terms of the correct localization (CorLoc) measure, which is the ratio of images with intersection over union with ground truth higher than 0.5. With the same number of strongly annotated images, additional weakly annotated images can be incorporated to give a 4.5\% point increase in CorLoc, from 80.00\% to 84.50\% (with 95\% confidence intervals 76.00\%--83.75\% and 81.00\%--88.00\%). The effects of different algorithmic details and varied amount of data are presented through ablative analysis. 

\end{abstract}

\begin{IEEEkeywords}
Breast ultrasound, convolutional neural networks, mass classification, mass localization, semi-supervised learning, weakly supervised learning.
\end{IEEEkeywords}

\section{Introduction}\label{sec:introduction}
\IEEEPARstart{B}{reast} cancer is the most common type of cancer among women worldwide, and approximately 1 in 8 (12\%) women in the US will develop invasive breast cancer during their lifetime~\cite{WCRFI13,stat16}. Consequently, annual breast cancer screening is crucial for its early detection. Ultrasound imaging is one of the modalities used in this screening. It is highly available, cost-effective, harmless, and has acceptable diagnostic performance. However, given that image acquisition and interpretation are often concurrently conducted by clinicians using traditional handheld ultrasound devices, it is likely to become more subjective than other imaging tests. Thus, efforts have been exerted to reduce subjectivity by standardizing ultrasound imaging procedures~\cite{birads13}.

\begin{figure}[t]
	\centering
	\includegraphics[width = 1\linewidth]{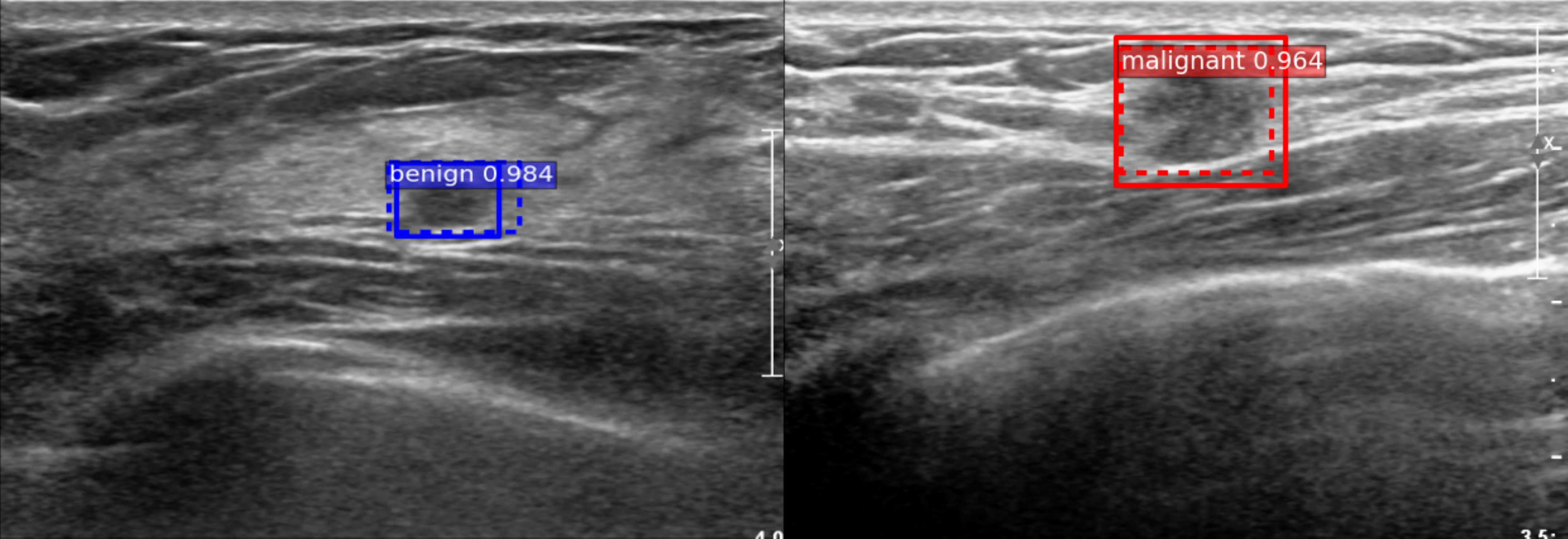}
	\caption{{ }Example of breast ultrasound images with masses. Bounding boxes with solid and dashed lines respectively represent ground truths (GT) and detections using the proposed method. Boxes are colored as blue (red) if the GT or predicted label is benign (malignant). Figure best viewed in color.}   
	\label{fig:example}
\end{figure}

Breast ultrasound (BUS) imaging aims to detect and classify abnormalities such as masses as either benign or malignant, as shown in Fig.~\ref{fig:example}. Most conventional methods are based on a sequential framework that comprises image preprocessing, region detection or segmentation, feature extraction, and classification~\cite{alvarenga07,huang08,shi10,minavathi12,gomez12}. These approaches require careful tweaking of each component, particularly because preceding processes affect subsequent processes. A summary of existing approaches for cancer detection and classification using BUS images can be found in \cite{cheng10}.

Conventional methods can be classified based on whether classification or region detection is emphasized. For methods emphasizing classification, manual or semi-automatic user interaction is often required to localize regions of interest (RoI). Compared to manual methods, semi-automatic methods only require sparse seeds, such as user clicks, for localization. For example, the method presented in~\cite{lee08} assumes manually segmented masses and uses those as input for subsequent processes. Most of these conventional approaches focus on developing effective features in distinguishing benign and malignant masses. Several manually designed geometric and echo features are used in~\cite{lee08}, and practical morphological features are defined from the contour of segmented lesions in~\cite{huang08}. A series of works also investigated gray-level co-occurrence matrix-based features~\cite{alvarenga07,gomez12,yang13}.

Other works mainly deal with fully automatic detection of the RoI. In \cite{drukker02}, radial gradient index filtering was used to detect initial points of interest, then lesion candidates were segmented by maximizing an average radial gradient for regions grown from the detected points and finally classified using Bayesian neural networks. In \cite{yap08}, hybrid filtering, multifractal processing, and thresholding segmentation were sequentially applied to detect all possible RoIs, and a rule-based approach was used to then identify the most important lesion. In \cite{jiang12}, a cascaded detector using Haar features within an AdaBoost learning framework were used to locate potential tumor locations, after which a support vector machine was utilized for refinement. In \cite{kisilev13}, lesions in BUS images were detected by pruning background edges from pre-detected lesion-specific edges. In \cite{pons14}, a generic object detection technique, namely deformable part models, was also adopted for this problem.

Recently, more approaches based on deep learning are being proposed~\cite{chen16,yap17,huynh16,cheng16,yap18}. Many of these methods are limited to either localizing target objects of interest~\cite{chen16,yap17} or classifying given RoIs into benign or malignant~\cite{huynh16,cheng16}, rather than conducting the two methods simultaneously. While the method recently proposed by Yap \emph{et al.}~\cite{yap18} performs both localization and classification simultaneously, it requires training data with lesion segmentation annotations since it is based on semantic image segmentation. This requirement may increase the burden of time and cost when compiling the training dataset.

Thus, we present a method for simultaneously localizing and classifying masses in BUS images. We train a convolutional neural network (CNN) for regression of bounding box positions and classification of masses, which can then be used to assign a per-image diagnostic label. Training such a model typically requires a strongly supervised image dataset, including the positions and labels of bounding boxes. While larger dataset size helps to avoid overfitting and maximize performance, considerable time and cost are required to obtain expert annotations. A dataset with weak annotations, e.g., image-level labels, which is often the case for BUS images, may be insufficient to train a model regardless of its size.

\begin{figure*}[t]
	\centering
	\includegraphics[width = 0.92\textwidth, height = 11.5cm]{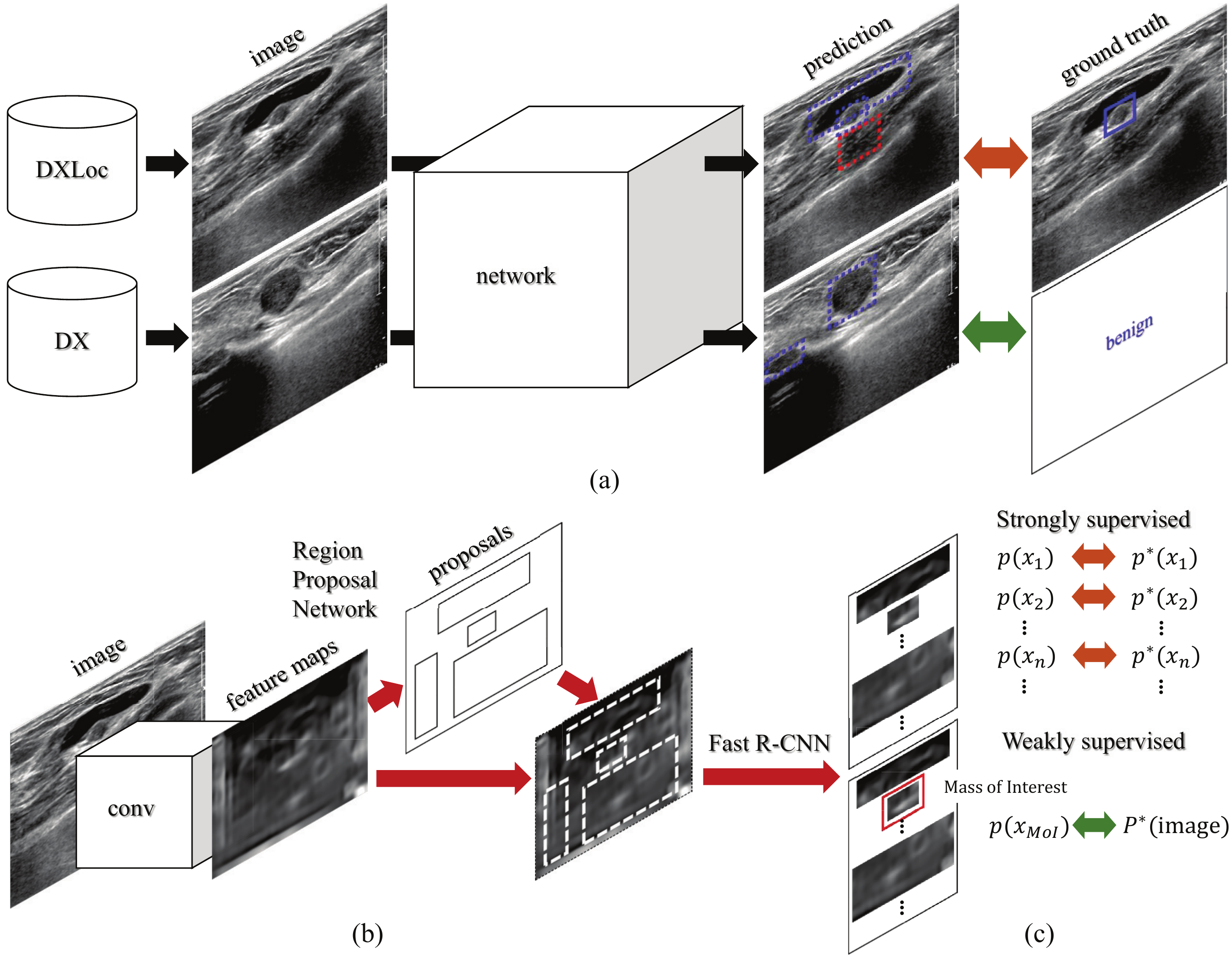}
	\caption{{ }Illustration of the proposed framework. (a) Images from two different data streams are forward-propagated into a shared network. (b) The Faster R-CNN~\cite{ren17} used for the ``network" of (a). The network is composed of the region proposal network (RPN) and Fast R-CNN~\cite{girshick15} with shared convolutional layers. This figure was previously presented in \cite{ren17} and is reprinted in this paper for the description of the Faster R-CNN. We also note that the proposed method is a general framework; hence, other supervised approaches can also be adopted. (c) An image-level loss is used for images from DX, whereas region-level losses are used for images from DXLoc. Refer to Subsections~\ref{subsec:sl} and~\ref{subsec:wsl} for details.}
	\label{fig:network}
\end{figure*}

Weakly supervised learning is a mechanism for datasets with noisy or sparse, i.e., weak, annotations. Multiple-instance learning (MIL) is a paradigm that defines the label relationship between a bag and its constituent multiple instances to learn more specific information from these weak labels. Most studies based on MIL have followed the work of Dietterich \emph{et al.}~\cite{dietterich97}, which was used for predicting drug activities. Recently, several pacesetting works have proposed the use of deep learning features in MIL tasks~\cite{xu14,song14}. Xu \emph{et al.}~\cite{xu14} studied colon cancer classification based on histopathology images. Song \emph{et al.}~\cite{song14} proposed a weakly supervised object detection method using deeply learned features. As an extension, Wu \emph{et al.}~\cite{wu15} incorporated learning deep representation into the MIL framework rather than merely using deep learning features as input. They redesigned a typical deep learning architecture to reflect the MIL assumption in image classification and annotation. Methods that incorporate MIL into the deep learning framework have also been proposed for medical imaging problems~\cite{yan15,shen16}. In the work of Yan \emph{et al.}\cite{yan15}, an approach to learn local discriminative regions per slice for body part recognition problem is proposed. In the work of Shen \emph{et al.}~\cite{shen16}, cancer malignancy of a patient is determined by aggregating nodule-level malignancies.

Approaches using datasets with different supervision levels have also been proposed~\cite{wu15wss,papandreou15,wang15,neverova17,souly17}. In Wu \emph{et al.}~\cite{wu15wss}, weakly labeled and unlabeled images are used for multi-label image annotation. Different losses are utilized to harness each image with varying characteristics. Similarly, weakly and strongly labeled images are used simultaneously for semantic image segmentation in~\cite{papandreou15,wang15,souly17}. Papandreou \emph{et al.} proposed expectation-maximization methods to train CNNs from weakly labeled images~\cite{papandreou15}. Wang \emph{et al.} proposed a method of two-stage training, first of a pixel-supervised CNN (PS-CNN) and then a collaborative-supervised CNN (CS-CNN)~\cite{wang15}. The CS-CNN is trained using image-level labels and pseudo pixel-wise labels generated by the PS-CNN, for weakly labeled images. Souly \emph{et al.}~\cite{souly17} extended typical generative adversarial networks~\cite{goodfellow14} to be applicable to semantic image segmentation and their networks were trained in a semi-supervised manner. In addition, Neverova \emph{et al.}~\cite{neverova17} also used weakly and strongly labeled images for hand pose estimation.

In this paper, we present a method to localize and classify masses from BUS images by training a CNN on a relatively small dataset with strong annotations and a large dataset with weak annotations in a hybrid manner. The proposed approach achieves a good balance between improved accuracy and reduced annotation cost. The method assumes a typical medical image setting where strong annotations are available for only a portion of images due to limited resources of physicians. Although the proposed framework exhibits similarities in terms of its aim with the methods presented in \cite{wu15wss,papandreou15,wang15,neverova17,souly17}, technical details differ significantly due to the different domains and objectives.

The main contributions of our work are the development of 1) a one-shot method for the concurrent localization and classification of masses present in BUS images and 2) a systematic weakly and semi-supervised training scenario for using a strongly annotated dataset, DXLoc and a weakly annotated dataset, DX, with appropriate training loss selection. Strong annotation comprises bounding box coordinates, denoted as Loc, and image-level diagnostic labels, denoted as DX, for all mass regions, while weak annotation comprises only image-level diagnostic labels. Two data streams are established during the training process, one for DXLoc and another for DX, while images from both sets are fed into a shared network, as shown in Fig.~\ref{fig:network}(a). Network weights are optimized by MIL loss in the DX data stream, and by losses computed based on given mass-level ground truth (GT) labels in the DXLoc data stream (Fig.~\ref{fig:network}(c)). Although various network models can be used, we provide the Faster R-CNN method~\cite{ren17} as an example in Fig.~\ref{fig:network}(b). The experiments show that the proposed method improves performance compared to methods using only DXLoc or DX. We believe the proposed method can be applied as a general framework to train a computer-aided diagnosis system for a wide variety of target organs and diseases.

\section{Methods}

\subsection{Datasets}\label{subsec:Datasets}

The proposed method is evaluated on two BUS datasets, which we refer to as SNUBH and UDIAT. All images can be classified by the amount of their label information. We refer to the set of images with only image-level diagnostic labels as DX, and images with both diagnostic labels and the bounding box of the mass of interest (MoI) as DXLoc. We note that we hereby use the term MoI, rather than RoI, to clarify that we are interested in the single most important mass. While the SNUBH dataset can be subdivided into SNUBH-DX and SNUBH-DXLoc, the UDIAT dataset all contains both annotations. Hence UDIAT-DXLoc is equivalent to UDIAT. The SNUBH-DX is only used in training, whereas the SNUBH-DXLoc and the UDIAT-DXLoc are further split by patients into training and test sets, namely, DXLoc-Tr and DXLoc-Ts.

The SNUBH dataset, collected from Seoul National University Bundang Hospital, comprises 5624 images from 3123 clinical cases and 2578 patients. The images were acquired using ultrasound systems from multiple vendors, including Philips (ATL HDI 5000, iU22), SuperSonic Imagine (Aixplorer), and Samsung Medison (RS80A), at 8 bit depth. All the images including masses have pathologically proven biopsy labels regarding benignancy or malignancy. The SNUBH-DXLoc subset comprises 1400 images (600 benign, 600 malignant, and 200 normal) in total. Only the bounding box and label of a single MoI is marked, whereas other probable masses are disregarded because operators intentionally focus only on the MoI. We thus explicitly draw background (non-mass) boxes as negative samples to exclude other probable masses. The SNUBH-DXLoc-Tr comprises 800 images (400 benign and 400 malignant), and SNUBH-DXLoc-Ts comprises 400 images (200 benign and 200 malignant). For completeness of comparative evaluations with previous methods, we also compiled a test dataset comprising only 200 normal images, separate from SNUBH-DXLoc-Ts. By contrast, the SNUBH-DX comprises 3291 benign and 933 malignant images. The ratio between benign and malignant in the SNUBH-DX follows natural statistics. Images with benign or malignant masses were collected from 2994 cases from 2449 patients, while normal images were collected from 129 cases from 129 patients. We note that the SNUBH dataset together with our code will be available online (\url{https://github.com/syshin1014/wssdl_bus}) for research purposes.

The UDIAT dataset, collected from the UDIAT Diagnostic Centre of the Parc Taul{\'i} Corporation, was originally used in\cite{yap17} and made available online (\url{http://goo.gl/SJmoti}) by the authors. Although this dataset was originally termed Dataset B in\cite{yap17}, we will refer to it as the UDIAT dataset for clarity in this paper. The dataset consists of 163 images from different patients. The images were acquired using a Siemens ACUSON Sequoia C512 system, and each of the images presented masses. Similar to the SNUBH dataset, all the images have diagnostic labels of masses present in the images. Consequently, 110 benign and 53 malignant images constitute the dataset. All the images also have respective mass segmentation annotations, which were required because masses are detected by conducting segmentation in\cite{yap17}. Refer to~\cite{yap17} for more details on the dataset. Since our proposed framework only requires bounding boxes and does not require segmentation annotations, which are more costly to obtain, we generated bounding box annotations from the segmentation boundaries. We randomly split the UDIAT dataset into the UDIAT-DXLoc-Tr set comprising 123 images (90 benign and 33 malignant) and the UDIAT-DXLoc-Ts set comprising 40 images (20 benign and 20 malignant).

\begin{figure}[t]
	\centering
	\subfloat{\includegraphics[width = 0.33\linewidth, height = 0.21\paperheight]{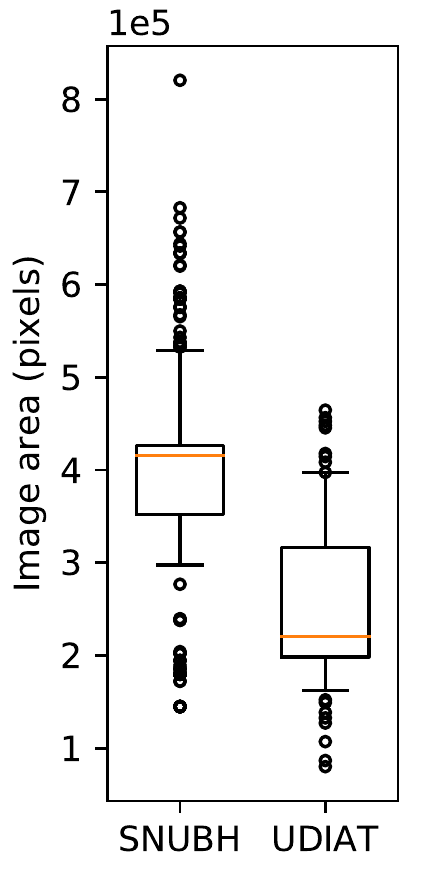}}
	\subfloat{\includegraphics[width = 0.33\linewidth, height = 0.21\paperheight]{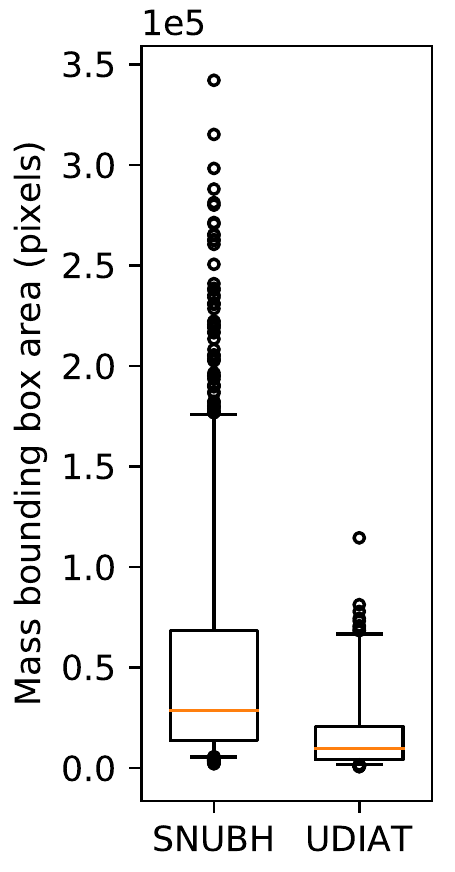}}
	\subfloat{\includegraphics[width = 0.33\linewidth, height = 0.21\paperheight]{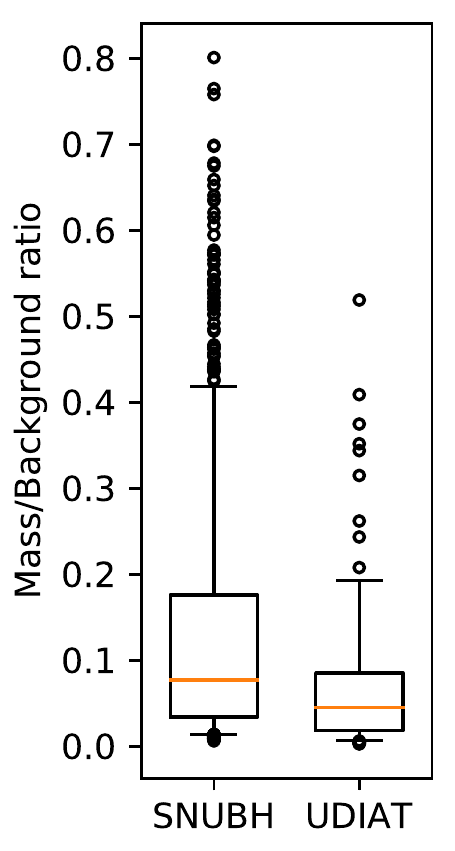}}
	\caption{{ }Comparison between two datasets. Box plots comparing (a) the whole image area, (b) the mass bounding box area, and (c) the ratio between the areas of the mass bounding box and the whole image.}
	\label{fig:stat}
\end{figure}

We summarize in Table~\ref{num_img_dataset} the numbers of images in the two datasets. In Fig.~\ref{fig:stat}, we compare the whole image area, the mass bounding box area, and the ratio between the areas of the mass bounding box and the whole image. For the SNUBH dataset, statistics of only the SNUBH-DXLoc are used. It shows several notable differences between the two datasets as follows: 1) The average size of the images in the SNUBH dataset is larger than that of the UDIAT dataset (Fig.~\ref{fig:stat}(a)). 2) The sizes of the masses (bounding boxes) in the SNUBH dataset exist in a wider range (Fig.~\ref{fig:stat}(b)(c)). We note that the finding wider range of the mass sizes of the SNUBH dataset, together with the fact that images in the SNUBH dataset were acquired using ultrasound systems from multiple vendors, implies a bigger diversity in the SNUBH dataset, which may imply higher difficulty in obtaining accurate results.

\begin{table}[tb]
   \caption{\textsc{Cardinality of SNUBH and UDIAT datasets}}
   \label{num_img_dataset}
   \centering
   \setlength{\tabcolsep}{3pt}
   \begin{tabular}{|c|c|c|c|c|c|c|c|c|}
      \hline
      \multicolumn{2}{|c|}{Dataset} & \multicolumn{4}{c|}{SNUBH} & \multicolumn{3}{c|}{UDIAT} \\
      \hline
      Role & Supervision & Nor. & Ben. & Mal. & Total & Ben. & Mal. & Total\\
      \hline
      \multirow{2}{*}{Training} & Strong (DXLoc) & 0 & 400 & 400 & 800 & 90 & 33 & 123\\
      \cline{2-9}
      & Weak (DX) & 0 & 3291 & 933 & 4224 & - & - & -\\
      \hline
      Test & Strong (DXLoc) & 200 & 200 & 200 & 600 & 20 & 20 & 40\\
      \hline
      \multicolumn{2}{|c|}{Total} & \multicolumn{4}{c|}{5624} & \multicolumn{3}{c|}{163} \\
      \hline
      \multicolumn{2}{|c|}{Cases/Patients} & \multicolumn{4}{c|}{3123/2578} & \multicolumn{3}{c|}{163/163} \\
      \hline
      \multicolumn{9}{p{240pt}}{Each dataset is further split by patients into subsets according to the role and the amount of supervision. Nor., Ben., and Mal. denote normal, benign, and malignant, respectively.}
   \end{tabular}
\end{table}

\subsection{Strongly Supervised Learning Using the DXLoc Subset}\label{subsec:sl}

Clinically, the final desired output is the image-level diagnostic label; during the process, a clinician inherently detects the MoI. Thus, we aim to perform MoI localization and classification jointly. To achieve this objective, we apply the Faster R-CNN method proposed in\cite{ren17} to our problem. Other supervised approaches can also be adopted.

The Faster R-CNN is composed of the region proposal network (RPN) and the Fast R-CNN detector~\cite{girshick15}, as shown in Fig.~\ref{fig:network}(b). The fully convolutional RPN generates rectangular \emph{region proposals}. The Fast R-CNN performs localization and classification on these proposals to detect objects of interests. The combined network is designed such that the RPN and the Fast R-CNN share the convolutional layers. This structure not only enables efficient region proposal generation and detection but also improves the precision of both tasks. 

The loss functions comprise four terms, which are the classification and regression losses $L^{RPN}_{cls}$, $L^{RPN}_{reg}$ and $L^{FRC}_{cls}$, $L^{FRC}_{reg}$ of the RPN and the Fast R-CNN, respectively. For all the terms, losses are based on the difference between the network outputs and the corresponding GT boxes defined by overlaps. In this study, several changes are necessary to adapt to our problem. For $L^{RPN}_{cls}$, proposals are defined as positive if the intersection over union (IoU) overlap with GT mass boxes is over 0.7, for either benign or malignant regions, for both the SNUBH and UDIAT datasets. Negative labels are differently assigned for each dataset. For the SNUBH dataset, a proposal is defined as negative if over 70\% of its area overlaps with one of the annotated background boxes, as described in Subsection~\ref{subsec:Datasets}. For the UDIAT dataset, we assign a negative label to a proposal if its IoU is lower than 0.3 for all GT mass boxes as in\cite{ren17}. Other proposals are disregarded. For $L^{FRC}_{cls}$, the probabilities for background, benign, and malignant classes are considered. The class weights are additionally considered for $L^{FRC}_{cls}$ when training the UDIAT dataset to address the class imbalance problem in the dataset. Regression losses are defined such that only positive proposals contribute to $L^{RPN}_{reg}$ and only detected benign or malignant boxes contribute to $L^{FRC}_{reg}$. In both regression losses, bounding box similarities are measured against the GT boxes with the highest IoU.

\subsection{Weakly Supervised Learning Using the DX Subset}\label{subsec:wsl}

The DX subset can be used to prevent overfitting and improve performance. Images from the DX set are fed into a network shared with DXLoc to produce region-level classification results, similar to the process in supervised learning. However, we incorporate a MIL scheme to define an appropriate loss function because no region-level GT label exists. In a MIL framework, a bag that comprises multiple instances is positive if at least one instance is positive and negative if all instances are negative. This assumption suggests that we can confidently label all instances in a negative bag as negative. Moreover, at least one instance in a positive bag will be positive. Compared with a similar MIL approach in~\cite{shen16}, one bag for each image in our problem regards all the detected mass regions as instances. If at least one region is classified as malignant, then that image is labeled as malignant. Thus, the per-image image-level loss $L_{ws}$ is defined as follows:

\begin{equation}
	\label{eq:MIL_cross_entropy}
	L_{ws}(I_{i}) = -\sum\limits_{l \in \mathcal L }P^{*}_{l}( I_{i})\log(P_{l}( I_{i})),
\end{equation}
where $L_{ws}( I_{i})$ is the cross entropy between GT label $P^{*}$ and prediction $P$ for the $i$th image $ I_{i}$. The class weights are omitted for brevity although cross entropies $L_{ws}( I_{i})$ multiplied by class weights are used in the experiments to address the class imbalance problem in the DX set.

The image-level label set ${\mathcal L} = \left\{ \emph{N}, \emph{B}, \emph{M} \right\}$ comprises normal (without any mass) \emph{N}, benign \emph{B}, and malignant \emph{M}. The image-level prediction $P$ is inherited from that of MoI $x_{MoI}$, where $x$ denotes a region, and $x_{MoI}$ denotes the MoI. Thus, $P_{l}$ is defined as
\begin{equation}
	\label{eq:img_level_prediction}
	P_{l}(I_{i}) = p_{l}(x_{MoI}), \forall l \in {\mathcal L}.
\end{equation}
The MoI can be selected based on several criteria. In particular, for images labeled \emph{B}, we test four different selection criteria, and each criterion selects the most benign \eqref{eq:mass_of_interest_benign}, malignant \eqref{eq:mass_of_interest_malignant}, discriminative \eqref{eq:mass_of_interest_discriminative}, or abnormal \eqref{eq:mass_of_interest_abnormal} region in the image, whereas the most malignant \eqref{eq:mass_of_interest_malignant} region is always selected as the MoI for images labeled \emph{M}:
\begin{equation}
	\label{eq:mass_of_interest_benign}
	x_{MoI} = \underset{x_{n} \in R(I_{i})}{\arg\max}~p_\emph{B}(x_n),
\end{equation}
\begin{equation}
	\label{eq:mass_of_interest_malignant}
	x_{MoI} = \underset{x_{n} \in R(I_{i})}{\arg\max}~p_\emph{M}(x_n),
\end{equation}
\begin{equation}
	\label{eq:mass_of_interest_discriminative}
	x_{MoI} = \underset{x_{n} \in R(I_{i})}{\arg\max}~\underset{l \in {\left\{\emph{B}, \emph{M}\right\}}}{\max}~p_{l}(x_n),
\end{equation}
\begin{equation}
	\label{eq:mass_of_interest_abnormal}
	x_{MoI} = \underset{x_{n} \in R(I_{i})}{\arg\min}~p_\emph{N}(x_n),
\end{equation}
where $R(I_i)$ is the set of detected regions for $I_i$, and $p_\emph{N}(x_n)$, $p_\emph{B}(x_n)$, and $p_\emph{M}(x_n)$ are the probabilities of normality, benignancy, and malignancy, respectively, for region $x_n$. Our definitions are based on the assumption that a clinician only focuses on the MoI. 

\subsection{Joint Weakly and Semi-Supervised Learning Using the DXLoc and DX Subsets}\label{subsec:wssl}

The components of strongly supervised learning and weakly supervised learning can be combined into a weakly and semi-supervised learning framework. We stream data from both DXLoc and DX to train the entire network composed of parameters $\theta = \theta_{conv} \cup \theta_{rpn} \cup \theta_{frcnn}$, comprising $\theta_{conv}$ of the shared convolutional layers, $\theta_{rpn}$ specific to the RPN, and $\theta_{frcnn}$ specific to the Fast R-CNN. We note that $\theta_{frcnn\_reg}$, related to the bounding box regression layer of the Fast R-CNN, are trained only by DXLoc. 

The network is trained by stochastic gradient descent (SGD) following the Faster R-CNN method~\cite{ren17} and convention of deep learning methodology. Within SGD, the two data streams can be used in two ways: (1) combining respective data instances into a single mini-batch (combined mini-batch), or (2) alternating between mini-batches of data from either stream (alternating mini-batch). For the combined mini-batch variant, the mini-batch in SGD comprises $b_{s}$ images from DXLoc and $b_{ws}$ images from DX, the loss functions regarding the strongly and weakly supervised data are summed as $L = L_{s} + {\alpha}L_{ws}$, where $L_{s} = L^{RPN}_{cls} + L^{RPN}_{reg} + L^{FRC}_{cls} + L^{FRC}_{reg}$ and learning rate $\eta$ is used to update $\theta$. For the alternating mini-batch variant, a mini-batch from DXLoc of size $b_{s}$ is first used with loss function $L_{s}$, with learning rate $\eta_{s}$ to update $\theta$. Then, a subsequent mini-batch from DX of size $b_{ws}$ is used with loss function ${\alpha}L_{ws}$, with learning rate $\eta_{ws}$ to update $\theta - \theta_{frcnn\_reg}$, the whole parameter set except for $\theta_{frcnn\_reg}$. In both variants, hyper-parameter $\alpha$ is defined to relatively scale the $L_{ws}$, the MIL loss \eqref{eq:MIL_cross_entropy}. The value of $\alpha$ is slowly increased from its given initial value to 1 during the training process, since the positive effect of MIL depends on the estimation accuracy of the network, which gradually increases during the network training.

\section{Results}

\subsection{Evaluation Details}

We use the ImageNet pre-trained VGG-16~\cite{simonyan14} model to initialize our network, and only fine-tune the layers $conv3\_1$ and up, as done in~\cite{ren17}. We reduce the sizes of the two fully connected layers of the Fast R-CNN to 512 and use a weight decay of 0.0005 to further prevent overfitting. For data augmentation, we apply horizontal flipping, random brightness, and contrast adjustment to the DXLoc-Tr and DX sets, and additional image-wise random rotation and central cropping to DX. Moreover, we use a simple Adam optimizer~\cite{kingma14} to reduce the number of hyperparameters for tuning. The parameters are fixed to $b_{s}=1$, $b_{ws}=2$, $\eta=\eta_{s}=\eta_{ws}=0.0005$, and $\alpha=0.01$. $b_{s}$ is fixed to 1 because multiple regions from an image actually constitute a mini-batch in supervised learning. For all methods, post-processes in the Faster R-CNN method~\cite{ren17} are also applied to generate the final detection outputs, including thresholding for initial detections and non-maximum suppression (NMS) based on class probabilities. All the settings above are equally adopted to all experiments unless otherwise noted.

\begin{table*}[t]
	\caption{\textsc{Ablation study showing algorithmic variants of the proposed method on the snubh dataset}}
	\label{ablation_res}
	\centering
	\setlength{\tabcolsep}{3pt}
	\begin{tabular}{|p{160pt}|p{170pt}|c|c|}
		\hline
		Aspect & Variant & CorLoc [\%] & 95\% Confidence Interval (CI) [\%] \\
		\hline
		\multirow{3}{*}{MIL definition of MoI for images labeled \emph{B} / \emph{M}}   & most benign region / most malignant region									& 79.50 & 75.75-83.50 \\
		& most discriminative region / most malignant region				& 80.50 & 76.50-84.25 \\
		& most abnormal region / most malignant region				& 82.00 & 78.25-85.75 \\
		\hline
		MIL scale factor 				& static value (0.5) 								& 82.25 & 78.25-86.00 \\
		\hline
		Combining weakly and strongly supervised data	& alternating mini-batches from each set 										& 82.50 & 78.75-86.25 \\
		\hline
		{\bf{Proposed}} & & {\bf 84.50} & 81.00-88.00 \\ MIL definition of MoI for images labeled \emph{B} / \emph{M} & most malignant region / most malignant region & & \\ MIL scale factor & gradual increase & & \\ Combining weakly and strongly supervised data & combined single mini-batch & & \\
		\hline
		\multicolumn{4}{p{500pt}}{Bootstrapping by randomly sampling test images with replacement is used to generate subset CorLoc values. From these sampled CorLoc values, 95\% confidence intervals (CI) are computed.}
	\end{tabular}
\end{table*}

The evaluations are conducted on the SNUBH-DXLoc-Ts or the UDIAT-DXLoc-Ts with the correct localization (CorLoc) measure~\cite{deselaers12} and the free-response receiver operating characteristic (FROC) curve~\cite{bunch78}. CorLoc is the percentage of images in which a method correctly localizes an object of the target class according to the PASCAL criterion (IoU $>$ 0.5)~\cite{everingham10}. Specifically, if the class probability of a detected region that has survived after NMS is larger than 0.5 for the target GT class and its IoU with the GT bounding box is over 0.5 it is determined to be correctly localized. For clarification in the context of medical images, we note that IoU is also known as the Jaccard Similarity Index (JSI). We believe CorLoc is more appropriate than mean average precision in our case because only the bounding box and the label of a single MoI is marked, whereas other probable masses are disregarded in the annotation process. 
The FROC curve is generated by varying the operating point, which is the threshold for the final class probabilities. That is, the class probability threshold for correct localization, which was defined as 0.5 for CorLoc, is varied to have values between 0 and 1.

To support the statistical significance of the performance of the proposed method, we also present the 95\% confidence interval (CI) of the CorLoc and the p-values obtained using a paired t-test for each comparable method. Both are computed based on bootstrapping, where a set of CorLoc's is generated by randomly sampling test images with replacement. 

\subsection{Experiments on the SNUBH Dataset}

\begin{table*}[t]
	\caption{\textsc{Ablation study of replacing the backbone cnn structure from vgg-16 \\to residual nets~\cite{he16} of varying depth on the snubh dataset}}
	\label{resnet_res}
	\centering
	\setlength{\tabcolsep}{3pt}
	\begin{tabular}{|c|c|c|c|c|c|c|c|c|c|c|c|c|c|c|c|}
		\hline
		\multicolumn{2}{|c|}{\multirow{2}{*}{Method}} & 
			\multicolumn{3}{c|}{\multirow{2}{*}{\cite{ren17}}} & 
			\multicolumn{3}{c|}{\multirow{2}{*}{\cite{ren17}+ResNet-34}}  & 
			\multicolumn{3}{c|}{\multirow{2}{*}{\cite{ren17}+ResNet-50}}  & 
			\multicolumn{3}{c|}{\multirow{2}{*}{\cite{ren17}+ResNet-101}}  & 
			Proposed+		& 
			\multirow{2}{*}{Proposed} \\
		\multicolumn{2}{|c|}{} &
			\multicolumn{3}{c|}{} & 
			\multicolumn{3}{c|}{} & 
			\multicolumn{3}{c|}{} &
			\multicolumn{3}{c|}{} &
			ResNet-101 & 
			\\
		\hline
		\multicolumn{2}{|c|}{CorLoc [\%]} & 
			64.25 & 77.25 & 80.00 & 
			69.25 & 77.75 & 78.25 & 
			64.50 & 75.25 & 79.25 & 
			61.50 & 74.25 & 81.25 & 
			83.25 & {\bf 84.50}\\
		\hline
		\multicolumn{2}{|c|}{\multirow{2}{*}{CI [\%]}} & 
			59.50- & 73.00- & 76.00- & 
			64.75- & 73.75- & 74.25- & 
			59.75- & 71.00- & 75.25- & 
			56.75- & 70.00- & 77.25- & 
			79.50- & 81.00-\\
		\multicolumn{2}{|c|}{} & 
			68.75 & 81.25 & 83.75 & 
			73.75 & 81.75 & 82.25 & 
			69.25 & 79.25 & 83.25 & 
			66.25 & 78.50 & 85.00 & 
			86.75 & 88.00\\
		\hline
		\# of training & 
		\#Strong  & 
			400 & 600 & 800 & 
			400 & 600 & 800 & 
			400 & 600 & 800 & 
			400 & 600 & 800 & 
			800	& 800 \\
		\cline{2-16} 
		images & 
		\#Weak  & 
			- & - & - & 
			- & - & - & 
			- & - & - & 
			- & - & - & 
			4224 & 4224 \\
		\hline
		\multicolumn{2}{|c|}{Level of supervision}	& \multicolumn{12}{c|}{Strong}				& \multicolumn{2}{c|}{Weak \& Semi} \\
		\hline
		\multicolumn{16}{p{500pt}}{Results with varying data size are also presented to show that increases in the network depth may be harmful if the amount of data is limited. Bootstrapping is used to sample subset CorLoc values from which 95\% confidence intervals (CI) are computed.}
	\end{tabular}
\end{table*}

\subsubsection{Ablation Study}

\paragraph{The Effect of Algorithmic Details}

The results of ablative analysis to investigate the influence of each algorithmic detail is presented in Table~\ref{ablation_res}. We first compare different MoI definitions for images labeled \emph{B}, as the most benign, malignant, discriminative, or abnormal regions. The MoI for images labeled \emph{M} is always the most malignant region. We found that defining the MoI as the most malignant region for images labeled \emph{B} as well exhibits the best result. This result is consistent with clinical practice where clinicians always focus on the most ``seemingly" malignant region as the MoI, and image-level diagnostic labels are determined based on the MoI label. We believe differences in MIL criteria between images with different labels introduce inconsistencies that cause poorer performance for the other MIL criteria. Next, the effectiveness of a gradually increasing scale factor for MIL loss, $\alpha$, is highlighted by the ``static value (0.5)" variant in Table~\ref{ablation_res}, where a static value is used for the scale factor from start to finish. In particular, 0.5 is used because DX has less precise information than DXLoc. The use of the gradually increasing scale factor helps prevent the network from converging to undesired local minima, which is confirmed by an increase of 2.25\% in performance. Lastly, we compare the results from SGD with combined DXLoc and DX mini-batches and from SGD with alternate DXLoc and DX mini-batches. We can see that the combined mini-batch variant exhibits a 2\% increase in performance by simultaneously probing DX and DXLoc while updating the parameters of the entire network.

\paragraph{The Effect of Network Structures}

We also show the applicability of the proposed method as a general framework by replacing VGG-16 with residual nets~\cite{he16} of varying layer depth, namely ResNet-34, ResNet-50, and ResNet-101. In this study, all the parameters remain unchanged regardless of network structure. Table~\ref{resnet_res} summarizes the results. Analogous to the result reported in \cite{ren17}, the performance of Faster R-CNN increased from 80.00\% with VGG-16 to 81.25\% with ResNet-101. The performance was also further increased when applying the proposed method of joint weakly and semi-supervised training with the ResNet-101. 

However, the performance achieved by applying the proposed method with the ResNet-101 decreases compared with that of the VGG-16. 
This led us to investigate further why residual nets did not perform better than VGG-16 for the proposed method. We thus conducted experiments with varying layer depth and varying data size (only with strongly supervised data to reduce the number of variants). Table~\ref{resnet_res} shows that while residual nets perform better with increasing layer depth for a strongly supervised training set of size 800, it actually performs worse with growing layer depth when that size is reduced to 600 or 400. These results show the current amount of limited data is not enough to fully realize the discriminative power of very deep residual nets. While we believe that deep residual nets will perform better with more training data, even weakly supervised, we could not verify this due to the limited amount of available data.

\paragraph{The Effect of Dataset Sizes}

\begin{figure}[t]
	\centering
	\includegraphics[width = 1\linewidth]{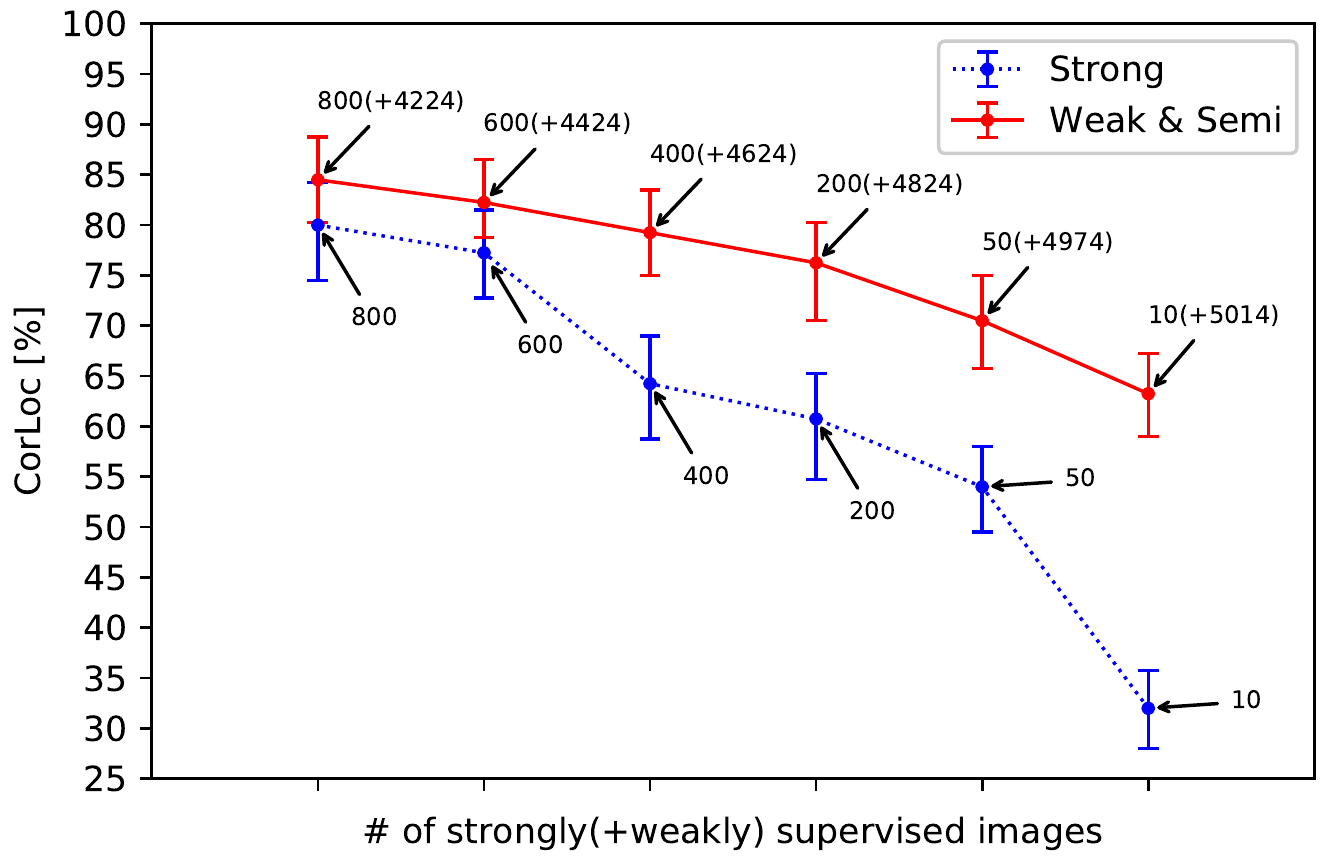}
	\caption{{ }The effect of sizes of training dataset with strong and weak supervision evaluated on the SNUBH dataset. CorLoc values on test set SNUBH-DXLoc-Ts of the proposed method, with both strongly supervised (DXLoc-Tr) and weakly supervised (DX) training data (in red), and with only strongly supervised training data (in blue). When only strongly supervised data are used, the proposed method is equivalent to the method of Ren \emph{et al.}~\cite{ren17}. The specific numbers of images of DXLoc-Tr and DX are presented for each marked point, and the 95\% confidence intervals of the measured CorLoc are also shown as bars.}
	\label{fig:graph}
\end{figure}

We tested the effect of the amount of data and supervision used in the training of our network by varying the sizes of the strongly and weakly supervised sets ($\#Strong/\#Weak$). Here, only a portion of the DXLoc-Tr set is used for strongly supervised training, whereas the rest are used for weakly supervised training, namely $800(+4224)$, $600(+4424)$, $400(+4624)$, $200(+4824)$, $50(+4974)$, $10(+5014)$, where $800(+4224)$ denotes 800 strongly supervised images with 4224 additional weakly supervised images and so on. A comparison between the results and the corresponding results using the same amounts of images only from the DXLoc-Tr set is represented in Fig.~\ref{fig:graph}. Here the effectiveness of the proposed method is clearly illustrated, which show better or comparable results with a considerably smaller amount of strongly supervised data complemented with weakly supervised data. Depending on the relative cost between the strong and weak annotations of a specific problem, an optimal ratio of strong and weak supervision may be determined to minimize annotation cost given a particular target performance. Moreover, by learning the DX set with the DXLoc-Tr set, the performance is relatively more robust when the size of the strongly supervised dataset DXLoc-Tr is small. This implies that weakly supervised data may have greater impact when the cost of achieving strongly supervised data is high. We can see in Fig.~\ref{fig:graph} that the proposed method performs much better when the number of images is extremely small, e.g., 50 and 10. 

\begin{table*}[bht]
	\caption{\textsc{Results of the proposed method with self-training on the snubh dataset}}
	\label{self_training_res}
	\centering
	\setlength{\tabcolsep}{3pt}
	\begin{tabular}{|c|c|c|c|c|c|c|c|}
		\hline
		\multicolumn{2}{|c|}{Model} & Initial & Retrained (50\%) & Retrained (75\%) & Initial & Retrained (50\%) & Retrained (75\%) \\
		\hline
		\multicolumn{2}{|c|}{CorLoc [\%]} & 70.50 & 82.25 & 80.75 & 63.25 & 81.00 & 79.25 \\
		\hline
		\multicolumn{2}{|c|}{CI [\%]} & 66.00--75.00 & 78.50--86.00 & 76.75--84.75 & 58.25--67.75 & 77.00--84.75 & 75.25--83.00 \\
		\hline
		\multirow{2}{*}{\# of training images} & \#Strong  	& 50 	  & 50+4974*0.5 & 50+4974*0.75& 10	   	& 10+5014*0.5  & 10+5014*0.75 \\
		\cline{2-8}
		 & \#Weak  	& 4974 	  & 4974*0.5 	& 4974*0.25	  & 5014 	& 5014*0.5 	  & 5014*0.25\\
		\hline
		\multicolumn{8}{p{450pt}}{The initial network is applied to the images in the weakly supervised set and images with the highest classification probabilities coincident with the image-level label are automatically selected and moved to the strongly supervised set, along with the most confident detection as the GT MoI. Retraining is conducted with the reconfigured strongly and weakly supervised sets. Bootstrapping is used to sample subset CorLoc values from which 95\% confidence intervals (CI) are computed.}
	\end{tabular}
\end{table*}

\begin{table*}[t]
	\caption{\textsc{Quantitative results of fully weakly supervised (weak), fully supervised (strong), and the proposed joint weakly and semi-supervised (weak and semi) methods on the SNUBH dataset}}
	\label{quan_res}
	\centering
	\setlength{\tabcolsep}{3pt}
	\begin{tabular}{|c|c|c|c|c|c|c|c|c|}
		\hline
		\multicolumn{2}{|c|}{\multirow{2}{*}{Method}} 	& \multirow{2}{*}{\cite{zhou16}} & \multirow{2}{*}{SS+MIL} 		& \multirow{2}{*}{\cite{ren17}} & \multicolumn{3}{c|}{\multirow{2}{*}{Proposed}} & Proposed+Self training \\
		\multicolumn{2}{|c|}{} 					&  &  		&  & \multicolumn{3}{c|}{} & (retrained 50\%)\\
		\hline
		\multicolumn{2}{|c|}{CorLoc [\%]} 	& 10.25 & 21.25 & 80.00 & {\bf 84.50} & 83.50 & 83.50 & 81.00\\
		\hline
		\multicolumn{2}{|c|}{CI [\%]} 		& 7.50-13.25 & 17.25--25.50 & 76.00--83.75 & 81.00--88.00 & 79.75--87.00 & 79.75--87.00 & -3.00--5.00 (difference with \cite{ren17})\\
		\hline
		\multicolumn{2}{|c|}{p-value} 		& $<$0.0001 & $<$0.0001 & $<$0.0001	& - & 0.0032 & 0.0018 & 0.6667 (with \cite{ren17})\\
		\hline
		\multirow{2}{*}{\# of training images} & \#Strong  & - 			 & - 		 	& 800 			& 800 			& 800	 & 800 & 10\\ 
		\cline{2-9}
		& \#Weak  & 5024 			 & 5024 		& - 			& 4224 			& 5024 	 & 2000 & 5014\\
		\hline
		\multicolumn{2}{|c|}{Level of supervision}	& \multicolumn{2}{c|}{Weak}	& Strong		& \multicolumn{4}{c|}{Weak \& Semi} \\ 			
		\hline
		\multicolumn{9}{p{450pt}}{Bootstrapping is used to sample subset CorLoc values from which 95\% confidence intervals (CI) and p-values for indicating statistical significance of improvement are computed. P-values are obtained using a paired t-test for each comparable method.}
	\end{tabular}
\end{table*}

\paragraph{The Effect of Using Self-training}

We further investigate the potential to reduce the size of strongly supervised data by using self-training~\cite{yarowsky95}. We first train an initial network with a given configuration of the DXLoc-Tr and DX sets. Then, we apply this initial network to the images in the DX set to obtain MoI detections. Images with the highest classification probabilities coincident with the image-level label are automatically selected and moved to the DXLoc-Tr set. Only one region, e.g., the most benign (or malignant) region in benign (or malignant) images, among all detections is used as a pseudo GT bounding box for the MoI since the image-level label only implies a single MoI. For the background, we automatically determine background boxes by finding boxes that do not overlap with any detections. Then a new network is trained with the new expanded DXLoc-Tr and reduced DX sets. Table~\ref{self_training_res} shows the results for an initial network trained with DXLoc-Tr and DX sets but reconfigured so that DXLoc-Tr has size 50 and 10. Results for the networks retrained after reclassifying 50\% of the DX set into the DXLoc-Tr set are also shown. It is shown that self-training improves the Corloc measures by 11.75\% and 17.75\% for DXLoc-Tr of size 50 and 10, respectively, compared to the initial networks. Here, we note that the reliability of the generated pseudo labels is most likely affected by the quality of the initial network model. Results for the retrained networks when reclassifying 75\% of the DX set show that reclassifying an excessive proportion may reduce the increase in performance.

\subsubsection{Comparison with Previous Methods}

We compare the proposed method with fully weakly supervised and fully supervised approaches on the SNUBH dataset. Fully weakly supervised methods include the methods of 1) Zhou \emph{et al.}~\cite{zhou16}, which can produce class-wise heat maps via ``class activation mapping," along with classification; and 2) MIL with region proposals generated via selective search~\cite{uijlings13} and our proposed loss, which is denoted as SS+MIL. In SS+MIL, the extracted region patches are resized to a fixed size and independently fed into a CNN. An image-level prediction is made by \eqref{eq:img_level_prediction}. The fully supervised method is that presented in \cite{ren17} and introduced in Subsection~\ref{subsec:sl}.

\paragraph{Quantitative Evaluation}

Table~\ref{quan_res} compares the proposed method with fully weakly supervised and fully supervised methods. We empirically found that training a fully weakly supervised model is non-trivial. We failed to train a competent mass detector due to the inferior image quality and complex patterns of BUS images. Our proposed joint weakly and semi-supervised approach clearly outperforms the other methods with statistical significance, supported by the confidence intervals and p-values.

In addition, we tested the effect of the amount of the weakly supervised set ($\#Weak$) while fixing that of the strongly supervised set ($\#Strong$) as follows: 1) All the images in DX and DXLoc-Tr are used for weakly supervised training ($800/5024$). 2) Only a portion of the DX set is used for weakly supervised learning ($800/2000$). The use of the maximum amount of data possibly provides the best result, except for the case of ($800/5024$). This result unexpectedly shows that applying the same data to supervised and weakly supervised training can have a negative effect. We conjecture that this phenomenon can occur when the loss from the weakly supervised MIL stream and the loss from the strongly supervised stream contradict each other for the same image. 

We can further assess the positive affect of additional training on weakly supervised images. It is shown in Table~\ref{quan_res} that the proposed method with self-training, trained only on 10 strongly and 5014 weakly supervised images, shows a slightly higher CorLoc than that of \cite{ren17} trained on 800 strongly supervised images. For a statistical comparison, we again use bootstrapping to sample subset CorLoc values of both settings to compute the 95\% CI of CorLoc differences as -3.00\%--5.00\%. The p-value for the null hypothesis was 0.6667.

\begin{figure}[t]
	\centering
	\includegraphics[width = 1\linewidth]{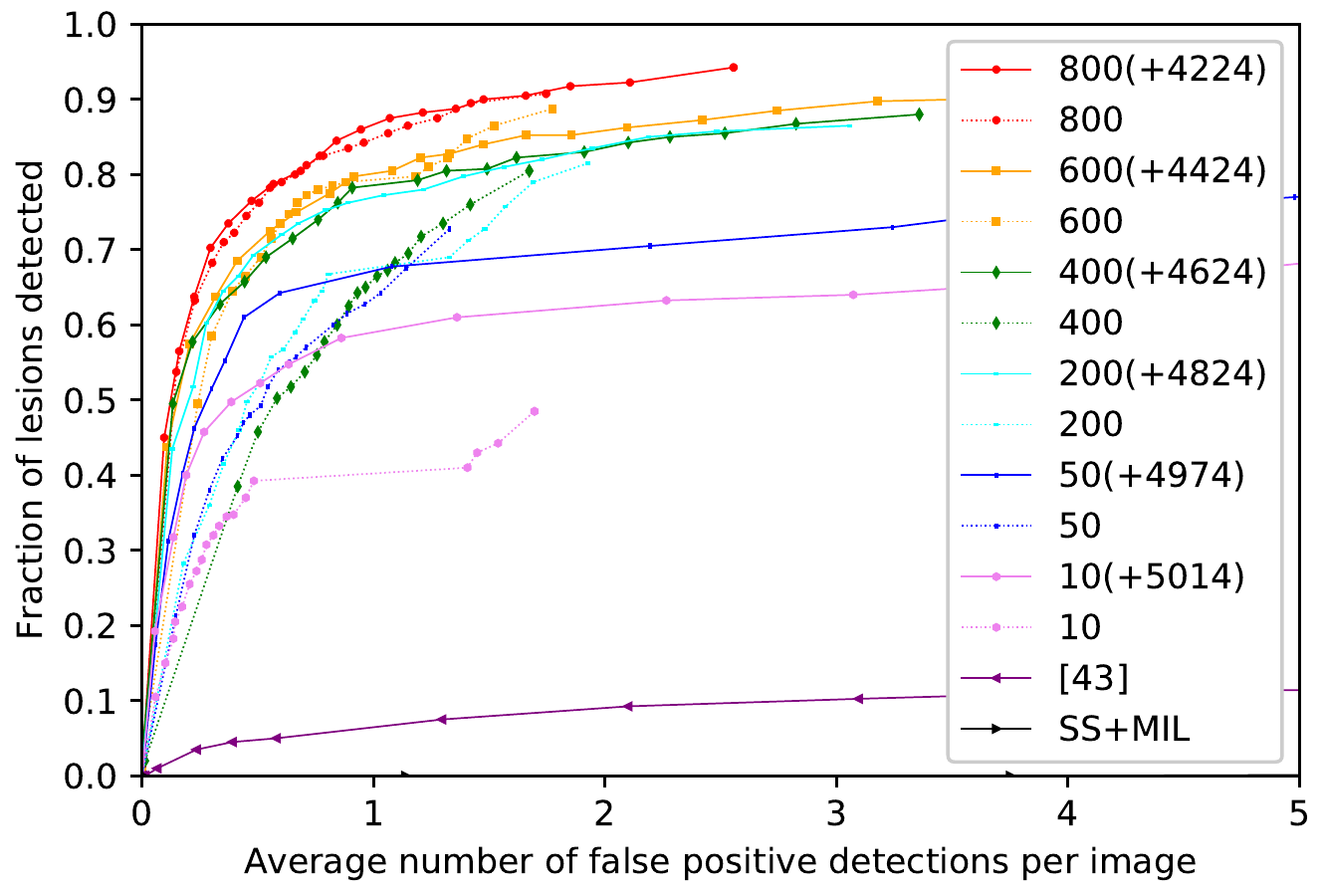}
	\caption{{ }FROC curves of the proposed method and comparable methods, corresponding to Table~\ref{quan_res}, on the SNUBH dataset. For the proposed method trained on both DXLoc-Tr (strongly supervised dataset) and DX (weakly supervised dataset), and trained only on DXLoc-Tr (where the proposed method becomes equivalent to \cite{ren17}), we show results for various dataset sizes. In the legend, 800(+4224) denotes 800 strongly and 4224 weakly supervised training images, and so on. The methods of Zhou \emph{et al.}~\cite{zhou16} and SS+MIL (MIL with region proposals generated via selective search~\cite{uijlings13}) are also shown.}
	\label{fig:froc_snubh}
\end{figure}

\begin{figure}[t]
	\centering
	\includegraphics[width = 1\linewidth]{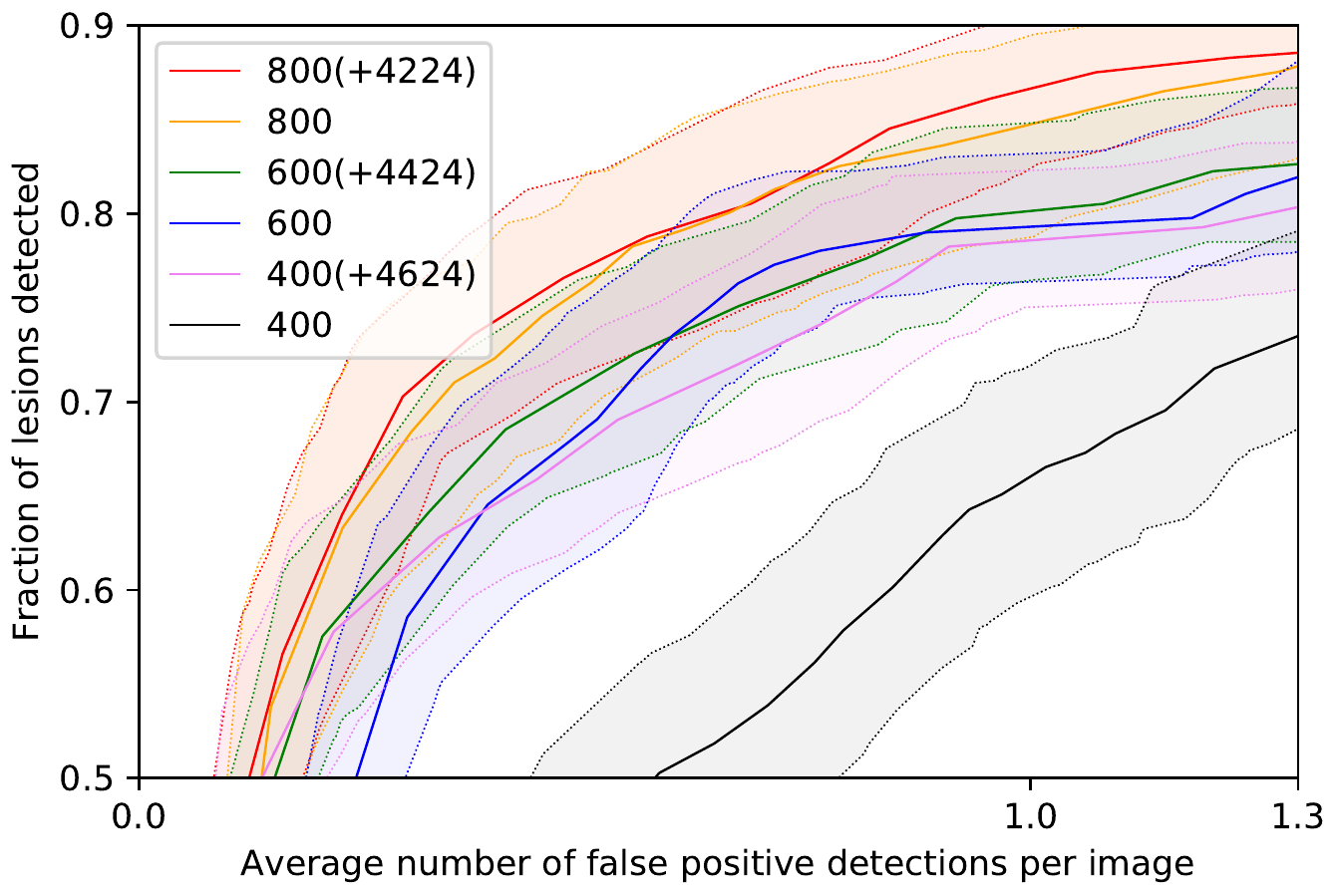}
	\caption{{ }Zoomed FROC curves with 95\% confidence interval corresponding to Fig.~\ref{fig:froc_snubh}. Only the topmost six curves are presented. Different colors are used for visibility. Bootstrapping by randomly sampling test images with replacement is used to generate multiple FROC curves for each model. From these, 95\% confidence intervals are computed.} 
	\label{fig:froc_snubh_ci}
\end{figure}

Fig.~\ref{fig:froc_snubh} shows the FROC curves for the methods included in Fig.~\ref{fig:graph} and Table~\ref{quan_res}. We can see that results from training only on strongly supervised data give final operating points that have fewer false positives (FP) and smaller fraction of detected lesions compared to the proposed training scheme. This indicates that the proposed weakly and semi-supervised training helps to detect more difficult MoIs at the expense of increased FPs. This is most likely beneficial in clinical applications of BUS, since it would be easier for clinicians to exclude FPs than to retroactively include false negatives. For further details of performance comparisons, we provide in Fig.~\ref{fig:froc_snubh_ci} the zoomed FROC curves with 95\% confidence interval corresponding to the six uppermost curves in Fig.~\ref{fig:froc_snubh}. 
%

Finally, we compared the performance of the proposed method on normal images. Since the proposed method can be applied to assist in screening for masses, we must consider the clinical environment where most images do not contain any at all and check if the improved performance of localization and classification comes at a cost of increased FPs on normal images. We use the aforementioned set of 200 normal images from 129 patients to obtain 2.07, 0.21, and 0.25 average number of FP detections per image for the method by Zhou \emph{et al.}~\cite{zhou16}, Ren \emph{et al.}~\cite{ren17}, and the proposed method, respectively. While the proposed method shows a 19.0\% increase in FP detections with a 5.63\% increase in CorLoc compared to the method of \cite{ren17}, we believe that the tradeoff is most likely worthwhile considering the relative difficulty of improving CorLoc for higher values and the absolute small number of FP detections.

\paragraph{Qualitative Evaluation}

\begin{figure*}[thb]
	\centering
	\includegraphics[width=1\linewidth, height = 0.78\paperheight]{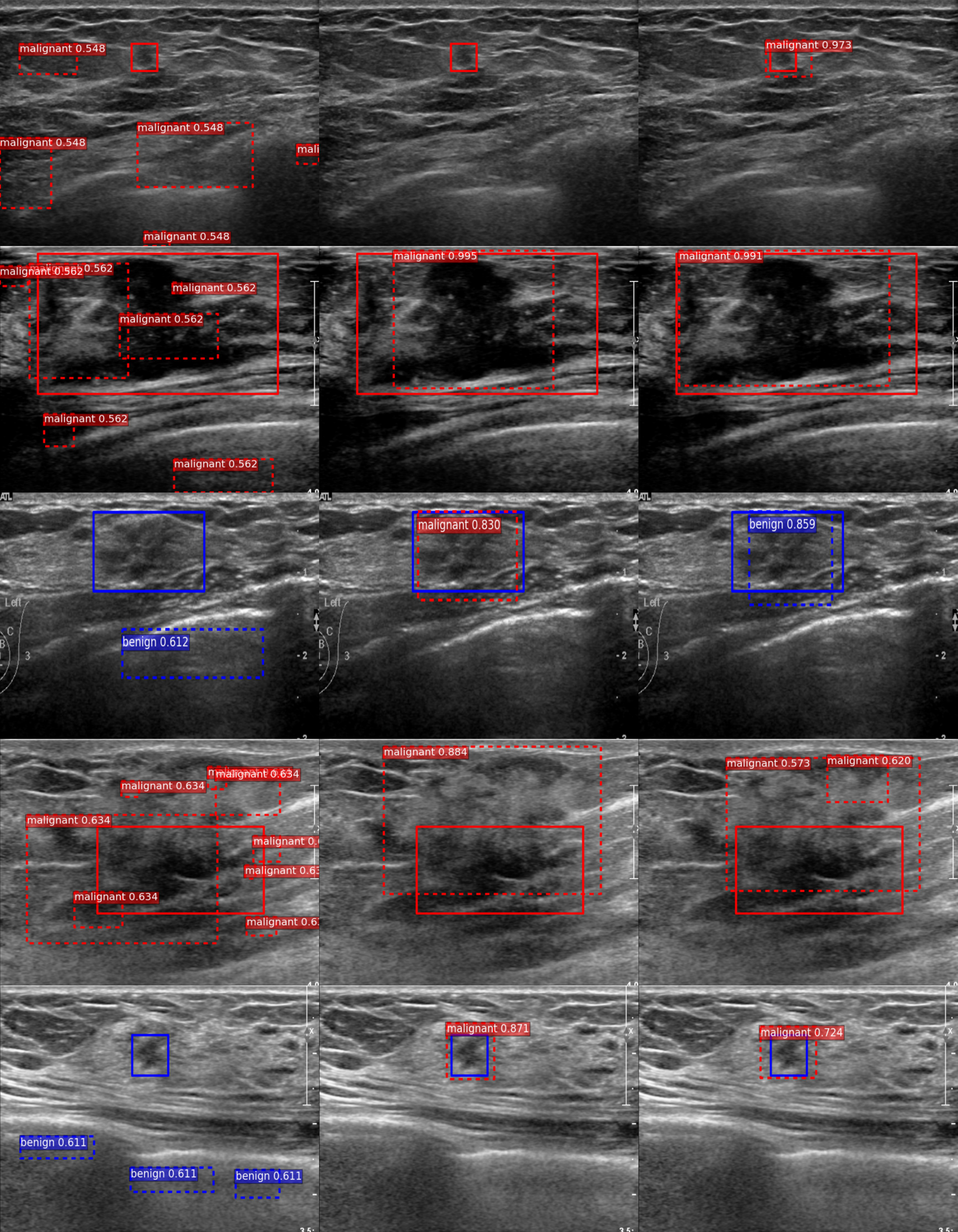}\\
	\begin{minipage}[b]{.33\linewidth}
		\centering{Weak~\cite{zhou16}}
	\end{minipage}%
	\begin{minipage}[b]{.33\linewidth}
		\centering{Strong~\cite{ren17}}
	\end{minipage}%
	\begin{minipage}[b]{.33\linewidth}
		\centering{Weak \& Semi (Proposed)}
	\end{minipage}%
	\caption{{ }Qualitative results on the SNUBH dataset. Each row shows different images. Each of the top three rows presents a case with various types of masses, which can be small, large, or unclear. The bottom two rows present failure cases where either localization or classification fails. Bounding boxes with solid and dashed lines respectively represent ground truths (GT) and detections using the proposed method. Boxes are colored as blue (red) if the GT or predicted label is benign (malignant). Figure best viewed in color.}
	\label{fig:qual_res_snubh}
\end{figure*}

Fig.~\ref{fig:qual_res_snubh} shows the qualitative results of the proposed method and those of the fully weakly supervised and fully supervised methods on the SNUBH dataset. From the first to the third row, the proposed method successfully detects and more precisely classifies various types of masses than the method presented in~\cite{ren17}.

Fig.~\ref{fig:qual_res_snubh} also provides representative failure cases in the SNUBH dataset. All the methods failed to detect a masse in the fourth row due to its unclear boundary. In the fifth example, the proposed method precisely localized but failed to correctly classify masses probably due to insufficient and confusing features. For example, a mass with an irregular shape and a nonparallel orientation in the example is likely to be seen as malignant from its image features.

\subsection{Experiments on the UDIAT Dataset}

\subsubsection{Comparison with Previous Methods}

\begin{table*}[tbh]
	\caption{\textsc{Quantitative results of fully supervised (strong) and the proposed \\joint weakly and semi-supervised (weak and semi) methods when training and testing across datasets}}
	\label{quan_res_comb}
	\centering
	\setlength{\tabcolsep}{3pt}
	\begin{tabular}{|c|c|c|c|c|c|}
		\hline
		\multicolumn{2}{|c|}{Method} & \cite{yap17}+\cite{cheng16}	& \cite{yap18} & \cite{ren17} & Proposed \\
		\hline
		\multirow{3}{*}{SNUBH-DXLoc-Ts} & CorLoc [\%] & 27.25 & 41.25 & 59.75 & {\bf 71.50} \\
		\cline{2-6}
		& CI [\%] & 23.00--31.75 & 36.50--46.00 & 55.00--64.25 & 67.25--76.00 \\
		\cline{2-6}
		& p-value & $<$0.0001 & $<$0.0001 & $<$0.0001 & - \\
		\hline
		\multirow{3}{*}{UDIAT-DXLoc-Ts} & CorLoc [\%] & 60.00 & 55.00 & 80.00 & {\bf 82.50} \\
		\cline{2-6}
		& CI [\%] & 55.25--64.75 & 40.00--70.00 & 76.00--83.75 & 78.50--86.00 \\
		\cline{2-6}
		& p-value & $<$0.0001 & $<$0.0001 & $<$0.0001 & - \\
		\hline
		\multicolumn{2}{|c|}{Level of supervision} & \multicolumn{3}{c|}{Strong} & Weak \& Semi \\
		\hline
		\multicolumn{6}{p{300pt}}{SNUBH-DX is used as the weakly supervised set only for the proposed method while the UDIAT-DXLoc-Tr is used as the strongly supervised set for all methods. The method(\cite{yap17}+\cite{cheng16}) comprises a method of localization by \cite{yap17} and classification by \cite{cheng16}. Bootstrapping is used to sample subset CorLoc values from which 95\% confidence intervals and p-values, by a paired t-test, for indicating statistical significance of improvement are computed.}
	\end{tabular}
\end{table*}

Comparisons are made with the recent approaches~\cite{yap17,cheng16,yap18}. We first construct a system, comparable to the proposed method, by cascading the two methods~\cite{yap17,cheng16} because each method only conducts localization~\cite{yap17} or classification~\cite{cheng16}. Masses are first localized and then classified in the cascaded system. Each network constituting the whole system was separately trained using the UDIAT-DXLoc-Tr. We note that training the network in\cite{yap17,yap18} requires segmentation annotations, which are available only for the UDIAT dataset. The FCN-AlexNet shows the best performance in\cite{yap17} in the absence of the FCN-VGG16, which showed the best performance in\cite{shelhamer17}. We thus use the FCN-VGG16 rather than the FCN-AlexNet in our experiment. The same modification on the network is applied for the method in\cite{yap18}. For these methods~\cite{yap17,cheng16,yap18}, we use our own implementations because the authors did not make theirs publicly available. We have carefully followed the paper descriptions for the best results.

\paragraph{Quantitative Evaluation}

Table~\ref{quan_res_comb} compares the proposed method with the cascaded system and the methods presented in~\cite{yap18,ren17} on the UDIAT-DXLoc-Ts. The SNUBH-DX is used as the weakly supervised set while the UDIAT-DXLoc-Tr is used as the strongly supervised set for all methods in common. Our proposed joint weakly and semi-supervised approach clearly outperforms the other methods on the UDIAT-DXLoc-Ts. We note that, the cascaded system and the method in \cite{yap18} require segmentation masks for training, which are very strong annotations compared to the mass bounding boxes and image-level diagnostic labels required by the proposed method. In addition, we applied the above networks, trained using the UDIAT-DXLoc-Tr, on the SNUBH-DXLoc-Ts. Our proposed approach performed reasonably well, whereas the other methods almost failed. We can see that by using the SNUBH-DX together with the small UDIAT-DXLoc-Tr for training, the performance is greatly boosted. We believe this result shows that our proposed method can be especially useful when strong annotation is hard to obtain. While training with only weakly annotated datasets may generally not be viable, if a common strongly annotated dataset is available, it can be jointly trained with various weakly annotated datasets in different environments, and successfully applied to test data from that environment. 

\begin{figure}[tb]
	\centering
	\includegraphics[width = 1\linewidth]{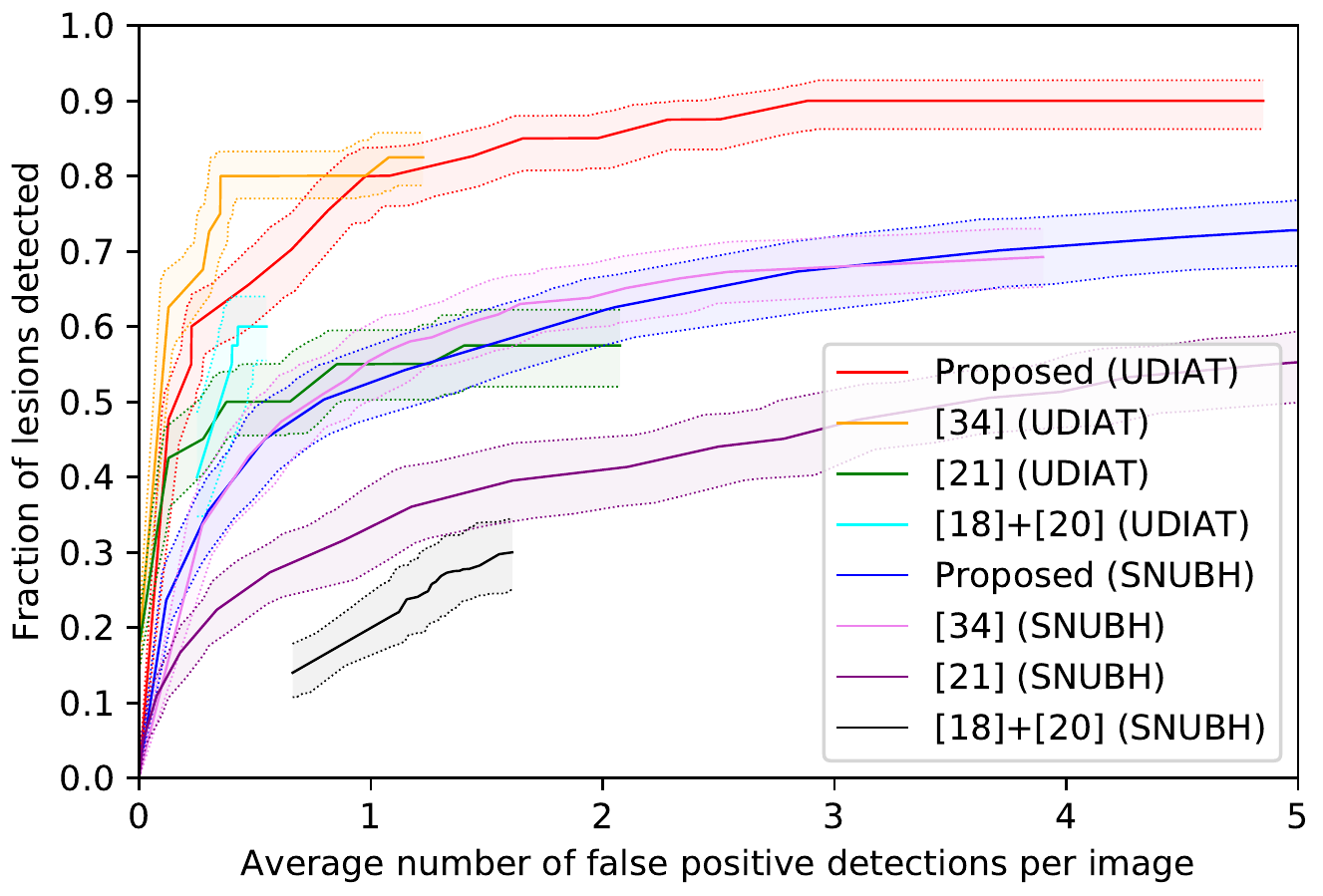}
	\caption{{ }FROC curves of the proposed joint weakly and semi-supervised method and comparable methods, corresponding to Table~\ref{quan_res_comb}, when training and testing across datasets. SNUBH-DX is used as the weakly supervised set only for the proposed method while the UDIAT-DXLoc-Tr is used as the strongly supervised set for all methods. Test sets are indicated in parentheses. Bootstrapping is used to generate multiple FROC curves from which 95\% confidence intervals are computed.} 
	\label{fig:froc_udiat}
\end{figure}

Fig.~\ref{fig:froc_udiat} shows the FROC curves for the all methods included in Table~\ref{quan_res_comb}. We can see that again, the proposed weakly and semi-supervised training helps to detect more difficult MoIs at the expense of increased FPs. 
%

\paragraph{Qualitative Evaluation}

\begin{figure*}[t]
	\centering
	\includegraphics[width=1\linewidth]{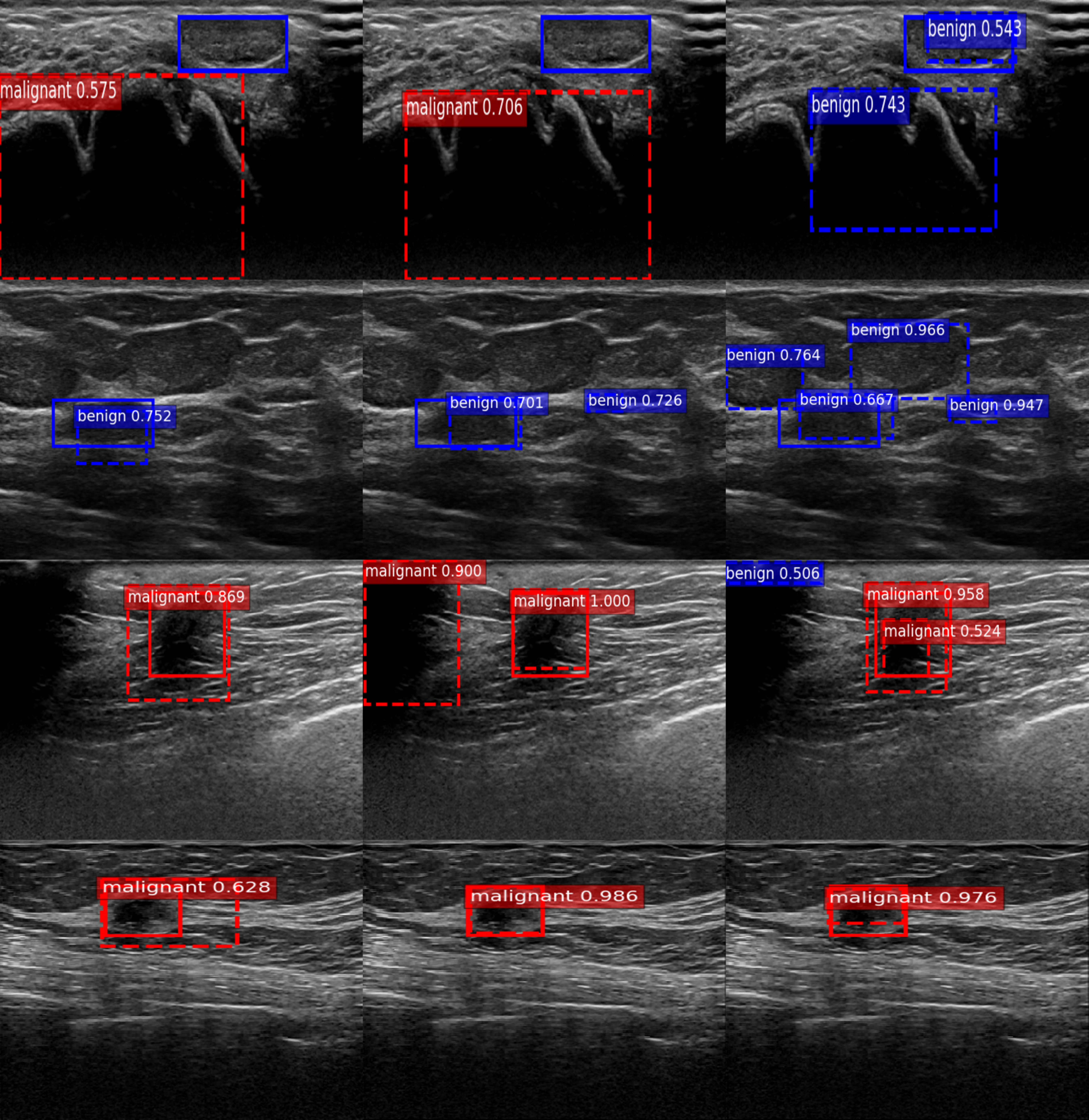}\\
	\begin{minipage}[b]{.33\linewidth}
		\centering{Strong~\cite{yap18}}
	\end{minipage}%
	\begin{minipage}[b]{.33\linewidth}
		\centering{Strong~\cite{ren17}}
	\end{minipage}%
	\begin{minipage}[b]{.33\linewidth}
		\centering{Weak \& Semi (Proposed)}
	\end{minipage}%
	\caption{{ }Qualitative results on the UDIAT-DXLoc-Ts. Each row shows different images. Bounding boxes with solid and dashed lines respectively represent ground truths (GT) and detections using the proposed method. Each two rows present benign and malignant cases. Boxes are colored as blue (red) if the GT or predicted label is benign (malignant). Figure best viewed in color.}
	\label{fig:qual_res_comb}
\end{figure*}

Fig.~\ref{fig:qual_res_comb} shows the qualitative results of the proposed method and those of the fully supervised methods on the UDIAT-DXLoc-Ts. The proposed method successfully detects and more precisely classifies both benign and malignant masses than the other methods.

\section{Conclusion}

We have proposed a systematic algorithm for weakly and semi-supervised learning, trained with a small number of strongly annotated data and a larger number of weakly annotated data. This method is applied to concurrently localize and classify masses present in BUS images. Comparative evaluation supports the effectiveness of the proposed method. Also, we have shown that combining the weakly annotated data in the training process will definitely improve the performance, moderately or greatly depending on the limitations of strongly supervised data. We have further investigated the following: the best particular version of the method, including details on algorithmic variants and network structure; the optimal configuration of weakly and strongly annotated data, considering required data annotation efforts and attainable accuracy; and the particular advantage of incorporating weakly annotated data in training. Our findings are: the optimal variant of the proposed method should be trained with mini-batches composed of both strongly and weakly annotated data, with gradually increasing MIL scaling, with MoI defined as the most malignant region, and with the VGG-16 net as the base network when data is limited; by using self-training, only an extremely small amount, e.g., under 100, of strongly annotated data can give comparable performance to when using a much larger amount, though it is better to increase strong annotations if the cost of annotation is low; and adding weakly annotated data will help to detect more true positives at the cost of increased false positives. 


For our future work, we plan to extend the proposed method to 3D automated BUS (ABUS). The ABUS has several advantages over hand-held US such as higher reproducibility and less required physician time for image acquisition. But image interpretation may be more time-consuming due to its higher dimensionality. 

We believe that although the proposed method is intended for masses in BUS images, it can also be applied as a general framework to train computer-aided detection and diagnosis systems to detect diverse lesion candidates before physicians manual inspection. Thus, the application of the proposed framework to a wide variety of image modalities, target organs, and diseases can also be our next step.



\begin{thebibliography}{10}
\providecommand{\url}[1]{#1}
\csname url@samestyle\endcsname
\providecommand{\newblock}{\relax}
\providecommand{\bibinfo}[2]{#2}
\providecommand{\BIBentrySTDinterwordspacing}{\spaceskip=0pt\relax}
\providecommand{\BIBentryALTinterwordstretchfactor}{4}
\providecommand{\BIBentryALTinterwordspacing}{\spaceskip=\fontdimen2\font plus
	\BIBentryALTinterwordstretchfactor\fontdimen3\font minus
	\fontdimen4\font\relax}
\providecommand{\BIBforeignlanguage}[2]{{%
		\expandafter\ifx\csname l@#1\endcsname\relax
		\typeout{** WARNING: IEEEtran.bst: No hyphenation pattern has been}%
		\typeout{** loaded for the language `#1'. Using the pattern for}%
		\typeout{** the default language instead.}%
		\else
		\language=\csname l@#1\endcsname
		\fi
		#2}}
\providecommand{\BIBdecl}{\relax}
\BIBdecl

\bibitem{WCRFI13}
\BIBentryALTinterwordspacing
(2013) {Breast cancer statistics}. [Online]. Available:
\url{http://www.wcrf.org/int/cancer-facts-figures/data-specific-cancers/breast-cancer-statistics}
\BIBentrySTDinterwordspacing

\bibitem{stat16}
\BIBentryALTinterwordspacing
(2016) {Risk of developing breast cancer}. [Online]. Available:
\url{http://www.breastcancer.org/symptoms/understand\_bc/risk/understanding}
\BIBentrySTDinterwordspacing

\bibitem{birads13}
\BIBentryALTinterwordspacing
American College of Radiology and C.~J.~D'Orsi, \emph{{ACR BI-RADS Atlas: Breast Imaging
		Reporting and Data System ; Mammography, Ultrasound, Magnetic Resonance
		Imaging, Follow-up and Outcome Monitoring, Data Dictionary}}.\hskip 1em plus
0.5em minus 0.4em\relax ACR, American College of Radiology, 2013. [Online].
Available: \url{https://books.google.co.kr/books?id=8Y6sAQAACAAJ}
\BIBentrySTDinterwordspacing

\bibitem{alvarenga07}
\BIBentryALTinterwordspacing
A.~V. Alvarenga, W.~C.~A. Pereira, A.~F.~C. Infantosi, and C.~M. Azevedo,
``{Complexity curve and grey level co-occurrence matrix in the texture
	evaluation of breast tumor on ultrasound images},'' \emph{Medical Physics},
vol.~34, no.~2, pp. 379--387, 2007. [Online]. Available:
\url{http://dx.doi.org/10.1118/1.2401039}
\BIBentrySTDinterwordspacing

\bibitem{huang08}
Y.-L. Huang, D.-R. Chen, Y.-R. Jiang, S.-J. Kuo, H.-K. Wu, and W.~Moon,
``{Computer-aided diagnosis using morphological features for classifying
	breast lesions on ultrasound},'' \emph{Ultrasound in Obstetrics \&
	Gynecology}, vol.~32, no.~4, pp. 565--572, 2008.

\bibitem{shi10}
\BIBentryALTinterwordspacing
X.~Shi, H.~D. Cheng, L.~Hu, W.~Ju, and J.~Tian, ``{Detection and Classification
	of Masses in Breast Ultrasound Images},'' \emph{Digit. Signal Process.},
vol.~20, no.~3, pp. 824--836, May 2010. [Online]. Available:
\url{https://doi.org/10.1016/j.dsp.2009.10.010}
\BIBentrySTDinterwordspacing

\bibitem{minavathi12}
Minavathi, M.~S, and M.~S. Dinesh, ``{Article: Classification of Mass in Breast
	Ultrasound Images using Image Processing Techniques},'' \emph{International
	Journal of Computer Applications}, vol.~42, no.~10, pp. 29--36, Mar 2012.

\bibitem{gomez12}
W.~Gomez, W.~C.~A. Pereira, and A.~F.~C. Infantosi, ``{Analysis of
	Co-Occurrence Texture Statistics as a Function of Gray-Level Quantization for
	Classifying Breast Ultrasound},'' \emph{IEEE Transactions on Medical
	Imaging}, vol.~31, no.~10, pp. 1889--1899, Oct 2012.

\bibitem{cheng10}
\BIBentryALTinterwordspacing
H.~D. Cheng, J.~Shan, W.~Ju, Y.~Guo, and L.~Zhang, ``{Automated Breast Cancer
	Detection and Classification Using Ultrasound Images: A Survey},''
\emph{Pattern Recognition}, vol.~43, no.~1, pp. 299--317, Jan 2010. [Online].
Available: \url{http://dx.doi.org/10.1016/j.patcog.2009.05.012}
\BIBentrySTDinterwordspacing

\bibitem{lee08}
\BIBentryALTinterwordspacing
G.~N. Lee, D.~Fukuoka, Y.~Ikedo, T.~Hara, H.~Fujita, E.~Takada, T.~Endo, and
T.~Morita, \emph{{Classification of Benign and Malignant Masses in Ultrasound
		Breast Image Based on Geometric and Echo Features}}.\hskip 1em plus 0.5em
minus 0.4em\relax Berlin, Heidelberg: Springer Berlin Heidelberg, 2008, pp.
433--439. [Online]. Available:
\url{https://doi.org/10.1007/978-3-540-70538-3\_60}
\BIBentrySTDinterwordspacing

\bibitem{yang13}
M.~C. Yang, W.~K. Moon, Y.~C.~F. Wang, M.~S. Bae, C.~S. Huang, J.~H. Chen, and
R.~F. Chang, ``{Robust Texture Analysis Using Multi-Resolution Gray-Scale
	Invariant Features for Breast Sonographic Tumor Diagnosis},'' \emph{IEEE
	Transactions on Medical Imaging}, vol.~32, no.~12, pp. 2262--2273, Dec 2013.

\bibitem{drukker02}
\BIBentryALTinterwordspacing
K.~Drukker, M.~L. Giger, K.~Horsch, M.~A. Kupinski, C.~J. Vyborny, and E.~B.
Mendelson, ``{Computerized lesion detection on breast ultrasound},''
\emph{Medical Physics}, vol.~29, no.~7, pp. 1438--1446, 2002. [Online].
Available: \url{http://dx.doi.org/10.1118/1.1485995}
\BIBentrySTDinterwordspacing

\bibitem{yap08}
M.~H. Yap, E.~A. Edirisinghe, and H.~E. Bez, ``{A novel algorithm for initial
	lesion detection in ultrasound breast images},'' \emph{Journal of Applied
	Clinical Medical Physics}, vol.~9, no.~4, pp. 181--199, 2008.

\bibitem{jiang12}
P.~Jiang, J.~Peng, G.~Zhang, E.~Cheng, V.~Megalooikonomou, and H.~Ling,
``{Learning-based automatic breast tumor detection and segmentation in
	ultrasound images},'' in \emph{Proceedings of IEEE International Symposium on
	Biomedical Imaging (ISBI)}, May 2012, pp. 1587--1590.

\bibitem{kisilev13}
P.~Kisilev, E.~Barkan, G.~Shakhnarovich, and A.~Tzadok, ``{Learning to detect lesion boundaries in breast ultrasound images},'' in  \emph{Proceedings of Workshop on Breast Image Analysisâin Conjunction with the 16th International Conference on Medical Image Computing and Computer-Assisted Intervention (MICCAI)}, Sep., 2013.

\bibitem{pons14}
\BIBentryALTinterwordspacing
G.~Pons, R.~Mart\'{i}, S.~Ganau, M.~Sent\'{i}s, and J.~Mart\'{i}, ``{Computerized
	Detection of Breast Lesions Using Deformable Part Models in Ultrasound
	Images},'' \emph{Ultrasound in Medicine \& Biology}, vol.~40, no.~9, pp.
2252--2264, 2014. [Online]. Available:
\url{http://www.sciencedirect.com/science/article/pii/S0301562914001537}
\BIBentrySTDinterwordspacing

\bibitem{chen16}
H.~Chen, Y.~Zheng, J.~H. Park, P.-A. Heng, and S.~K. Zhou, ``{Iterative
	Multi-domain Regularized Deep Learning for Anatomical Structure Detection and
	Segmentation from Ultrasound Images},'' in \emph{Proceedings of International Conference on
	Medical Image Computing and Computer-Assisted Intervention (MICCAI)}.\hskip 1em plus
0.5em minus 0.4em\relax Springer, 2016, pp. 487--495.

\bibitem{yap17}
M.~H.~Yap, G.~Pons, J.~Mart\'{i}, S.~Ganau, M.~Sent\'{i}s, R.~Zwiggelaar, A.~K.
Davison, and R.~Mart\'{i}, ``{Automated Breast Ultrasound Lesions Detection
	using Convolutional Neural Networks},'' \emph{IEEE Journal of Biomedical and
	Health Informatics}, vol.~PP, no.~99, pp. 1--1, 2017.

\bibitem{huynh16}
\BIBentryALTinterwordspacing
B.~Huynh, K.~Drukker, and M.~Giger, ``{MO-DE-207B-06: Computer-Aided Diagnosis
	of Breast Ultrasound Images Using Transfer Learning From Deep Convolutional
	Neural Networks},'' \emph{Medical Physics}, vol.~43, no. 6Part30, pp.
3705--3705, 2016. [Online]. Available:
\url{http://dx.doi.org/10.1118/1.4957255}
\BIBentrySTDinterwordspacing

\bibitem{cheng16}
J.~Cheng, D.~Ni, Y.~Chou, J.~Qin, C.~Tiu, Y.~Chang, C.~Huang, D.~Shen, and
C.~Chen, ``{Computer-Aided Diagnosis with Deep Learning Architecture:
	Applications to Breast Lesions in US Images and Pulmonary Nodules in CT
	Scans},'' \emph{Scientific Reports}, vol.~6, Apr 2016.

\bibitem{yap18}
M.~H.~Yap, M.~Goyal, F.~Osman, E.~Ahmad, R.~Mart\'{i}, E.~Denton, A.~Juette, R.~Zwiggelaar, ``{End-to-end breast ultrasound lesions recognition with a deep learning approach},''  in \emph{Proceedings of SPIE 10578, Medical Imaging}, vol.~1057819,  Mar 2018. [Online]. Available: \url{https://doi.org/10.1117/12.2293498}


\bibitem{dietterich97}
\BIBentryALTinterwordspacing
T.~G. Dietterich, R.~H. Lathrop, and T.~Lozano-P\'{e}rez, ``{Solving the multiple
	instance problem with axis-parallel rectangles},'' \emph{Artificial
	Intelligence}, vol.~89, no.~1, pp. 31 -- 71, 1997. [Online]. Available:
\url{http://www.sciencedirect.com/science/article/pii/S0004370296000343}
\BIBentrySTDinterwordspacing

\bibitem{xu14}
Y.~Xu, T.~Mo, Q.~Feng, P.~Zhong, M.~Lai, and E.~I.~C. Chang, ``{Deep learning
	of feature representation with multiple instance learning for medical image
	analysis},'' in \emph{Proceedings of IEEE International Conference on Acoustics, Speech
	and Signal Processing (ICASSP)}, May 2014, pp. 1626--1630.

\bibitem{song14}
\BIBentryALTinterwordspacing
H.~O. Song, Y.~J. Lee, S.~Jegelka, and T.~Darrell, ``{Weakly-supervised
	Discovery of Visual Pattern Configurations},'' in \emph{Proceedings of 27th International Conference on Neural Information Processing Systems}, ser.
NIPS'14.\hskip 1em plus 0.5em minus 0.4em\relax Cambridge, MA, USA: MIT
Press, 2014, pp. 1637--1645. [Online]. Available:
\url{http://dl.acm.org/citation.cfm?id=2968826.2969009}
\BIBentrySTDinterwordspacing

\bibitem{wu15}
J.~Wu, Y.~Yu, C.~Huang, and K.~Yu, ``{Deep multiple instance learning for image
	classification and auto-annotation},'' in \emph{Proceedings of IEEE Conference on
	Computer Vision and Pattern Recognition (CVPR)}, Jun 2015, pp. 3460--3469.

\bibitem{yan15}
\BIBentryALTinterwordspacing
Z.~Yan, Y.~Zhan, Z.~Peng, S.~Liao, Y.~Shinagawa, D.~N. Metaxas, and X.~S. Zhou,
\emph{{Bodypart Recognition Using Multi-stage Deep Learning}}.\hskip 1em plus
0.5em minus 0.4em\relax Cham: Springer International Publishing, 2015, pp.
449--461. [Online]. Available:
\url{https://doi.org/10.1007/978-3-319-19992-4\_35}
\BIBentrySTDinterwordspacing

\bibitem{shen16}
\BIBentryALTinterwordspacing
W.~Shen, M.~Zhou, F.~Yang, D.~Dong, C.~Yang, Y.~Zang, and J.~Tian,
\emph{{Learning from Experts: Developing Transferable Deep Features for
		Patient-Level Lung Cancer Prediction}}.\hskip 1em plus 0.5em minus
0.4em\relax Cham: Springer International Publishing, 2016, pp. 124--131.
[Online]. Available: \url{https://doi.org/10.1007/978-3-319-46723-8\_15}
\BIBentrySTDinterwordspacing

\bibitem{wu15wss}
F.~Wu, Z.~Wang, Z.~Zhang, Y.~Yang, J.~Luo, W.~Zhu, and Y.~Zhuang, ``{Weakly
	Semi-Supervised Deep Learning for Multi-Label Image Annotation},'' \emph{IEEE
	Transactions on Big Data}, vol.~1, no.~3, pp. 109--122, Sep 2015.

\bibitem{papandreou15}
G.~Papandreou, L.~C. Chen, K.~P. Murphy, and A.~L. Yuille, ``{Weakly-and
	Semi-Supervised Learning of a Deep Convolutional Network for Semantic Image
	Segmentation},'' in \emph{Proceedings of IEEE International Conference on Computer
	Vision (ICCV)}, Dec 2015, pp. 1742--1750.

\bibitem{wang15}
\BIBentryALTinterwordspacing
Y.~Wang, J.~Liu, Y.~Li, and H.~Lu, ``{Semi- and Weakly- Supervised Semantic
	Segmentation with Deep Convolutional Neural Networks},'' in \emph{Proceedings
	of ACM International Conference on Multimedia}, ser. MM '15.\hskip
1em plus 0.5em minus 0.4em\relax New York, NY, USA: ACM, 2015, pp.
1223--1226. [Online]. Available:
\url{http://doi.acm.org/10.1145/2733373.2806322}
\BIBentrySTDinterwordspacing

\bibitem{neverova17}
\BIBentryALTinterwordspacing
N.~Neverova, C.~Wolf, F.~Nebout, and G.~W. Taylor, ``{Hand pose estimation
	through semi-supervised and weakly-supervised learning},'' \emph{Computer
	Vision and Image Understanding}, 2017. [Online]. Available:
\url{http://www.sciencedirect.com/science/article/pii/S1077314217301686}
\BIBentrySTDinterwordspacing

\bibitem{souly17}
N.~Souly, C.~Spampinato, and M.~Shah, ``{Semi Supervised Semantic Segmentation
	Using Generative Adversarial Network},'' in \emph{Proceedings of IEEE International
	Conference on Computer Vision (ICCV)}, Oct 2017.

\bibitem{goodfellow14}
\BIBentryALTinterwordspacing
I.~Goodfellow, J.~Pouget-Abadie, M.~Mirza, B.~Xu, D.~Warde-Farley, S.~Ozair,
A.~Courville, and Y.~Bengio, ``{Generative Adversarial Nets},'' in
\emph{Advances in Neural Information Processing Systems 27}, Z.~Ghahramani,
M.~Welling, C.~Cortes, N.~D. Lawrence, and K.~Q. Weinberger, Eds.\hskip 1em
plus 0.5em minus 0.4em\relax Curran Associates, Inc., 2014, pp. 2672--2680.
[Online]. Available:
\url{http://papers.nips.cc/paper/5423-generative-adversarial-nets.pdf}
\BIBentrySTDinterwordspacing

\bibitem{ren17}
S.~Ren, K.~He, R.~Girshick, and J.~Sun, ``{Faster R-CNN: Towards Real-Time
	Object Detection with Region Proposal Networks},'' \emph{IEEE Transactions on
	Pattern Analysis and Machine Intelligence}, vol.~39, no.~6, pp. 1137--1149,
Jun 2017.

\bibitem{girshick15}
R.~Girshick, ``{Fast R-CNN},'' in \emph{Proceedings of IEEE International Conference on
	Computer Vision (ICCV)}, Dec 2015, pp. 1440--1448.

\bibitem{simonyan14}
K.~Simonyan and A.~Zisserman, ``{Very Deep Convolutional Networks for
	Large-Scale Image Recognition},'' \emph{CoRR}, vol. abs/1409.1556, 2014.

\bibitem{kingma14}
\BIBentryALTinterwordspacing
D.~P. Kingma and J.~Ba, ``{Adam: {A} Method for Stochastic Optimization},''
\emph{CoRR}, vol. abs/1412.6980, 2014. [Online]. Available:
\url{http://arxiv.org/abs/1412.6980}
\BIBentrySTDinterwordspacing

\bibitem{deselaers12}
\BIBentryALTinterwordspacing
T.~Deselaers, B.~Alexe, and V.~Ferrari, ``{Weakly Supervised Localization and
	Learning with Generic Knowledge},'' \emph{International Journal of Computer
	Vision}, vol. 100, no.~3, pp. 275--293, Dec 2012. [Online]. Available:
\url{https://doi.org/10.1007/s11263-012-0538-3}
\BIBentrySTDinterwordspacing

\bibitem{bunch78}
\BIBentryALTinterwordspacing
P.~Bunch, J.~Hamilton, G.~Sanderson, and A.~Simmons, ``{Free response approach
   to measurement and characterization of radiographic observer performance},''
\emph{American Journal of Roentgenology}, vol. 130, no.~2, pp. 382--382,
1978.
\BIBentrySTDinterwordspacing

\bibitem{everingham10}
\BIBentryALTinterwordspacing
M.~Everingham, L.~Gool, C.~K. Williams, J.~Winn, and A.~Zisserman, ``{The
	Pascal Visual Object Classes (VOC) Challenge},'' \emph{International Journal of Computer Vision}, vol.~88, no.~2, pp. 303--338, Jun 2010. [Online]. Available:
\url{http://dx.doi.org/10.1007/s11263-009-0275-4}
\BIBentrySTDinterwordspacing

\bibitem{he16}
K.~He, X.~Zhang, S.~Ren, and J.~Sun, ``{Deep Residual Learning for Image
	Recognition},'' in \emph{Proceedings of IEEE Conference on Computer Vision and Pattern
	Recognition (CVPR)}, Jun 2016, pp. 770--778.

\bibitem{yarowsky95}
\BIBentryALTinterwordspacing
D.~Yarowsky, ``{Unsupervised Word Sense Disambiguation Rivaling Supervised
	Methods},'' in \emph{Proceedings of the 33rd Annual Meeting on Association
	for Computational Linguistics}, ser. ACL '95.\hskip 1em plus 0.5em minus
0.4em\relax Stroudsburg, PA, USA: Association for Computational Linguistics,
1995, pp. 189--196. [Online]. Available:
\url{https://doi.org/10.3115/981658.981684}
\BIBentrySTDinterwordspacing

\bibitem{zhou16}
B.~Zhou, A.~Khosla, A.~Lapedriza, A.~Oliva, and A.~Torralba, ``{Learning Deep
	Features for Discriminative Localization},'' in \emph{Proceedings of IEEE Conference on
	Computer Vision and Pattern Recognition (CVPR)}, Jun 2016, pp. 2921--2929.

\bibitem{uijlings13}
\BIBentryALTinterwordspacing
J.~R.~R. Uijlings, K.~E.~A. van~de Sande, T.~Gevers, and A.~W.~M. Smeulders,
``{Selective Search for Object Recognition},'' \emph{International Journal of
	Computer Vision}, vol. 104, no.~2, pp. 154--171, Sep 2013. [Online].
Available: \url{https://doi.org/10.1007/s11263-013-0620-5}
\BIBentrySTDinterwordspacing

\bibitem{shelhamer17}
E.~Shelhamer, J.~Long, and T.~Darrell, ``{Fully Convolutional Networks for
	Semantic Segmentation},'' \emph{IEEE Transactions on Pattern Analysis and
	Machine Intelligence}, vol.~39, no.~4, pp. 640--651, Apr 2017.
	
\end{thebibliography}
\end{document}